\documentclass[english]{IEEEtran}
\usepackage[T1]{fontenc}
\usepackage[latin9]{inputenc}
\usepackage{color}
\usepackage{verbatim}
\usepackage{float}
\usepackage{amsmath}
\usepackage{amsthm}
\usepackage{amssymb}
\usepackage{stmaryrd}
\usepackage{stackrel}
\usepackage{graphicx}
\usepackage{wasysym}
\usepackage{mathtools}
\usepackage{hyperref}
\usepackage{url}   
\usepackage{booktabs} 
\usepackage{nicefrac} 
\usepackage{authblk}
\usepackage{tikzpagenodes}
\usepackage{paralist}
\makeatletter

\floatstyle{ruled}
\newfloat{algorithm}{tbp}{loa}
\providecommand{\algorithmname}{Algorithm}
\floatname{algorithm}{\protect\algorithmname}

\theoremstyle{plain}
\newtheorem{thm}{\protect\theoremname}
\theoremstyle{definition}
\newtheorem{defn}{\protect\definitionname}
\theoremstyle{definition}
\newtheorem{problem}{\protect\problemname}
\theoremstyle{plain}
\newtheorem{lem}{\protect\lemmaname}
\theoremstyle{definition}
\newtheorem{example}{\protect\examplename}
\theoremstyle{remark}
\newtheorem{rem}{\protect\remarkname}
\theoremstyle{plain}

\usepackage{cite}
\usepackage{algpseudocode}
\usepackage{setspace}
\IEEEoverridecommandlockouts

\makeatother

\usepackage{babel}
\providecommand{\corollaryname}{Corollary}
\providecommand{\definitionname}{Definition}
\providecommand{\examplename}{Example}
\providecommand{\lemmaname}{Lemma}
\providecommand{\problemname}{Problem}
\providecommand{\remarkname}{Remark}
\providecommand{\theoremname}{Theorem}

\newcommand{\UntilOp}{\mathcal{U}}

\newcommand{\AP}{{AP}}

\SetSymbolFont{stmry}{bold}{U}{stmry}{m}{n}

\usepackage{cite}
\usepackage{algpseudocode}
\usepackage{setspace}
\usepackage{subfig}
\usepackage{color}
\usepackage{tikz}
\tikzset{>=latex}
\usetikzlibrary{fit,automata,positioning,angles,quotes}

\def\BibTeX{{\rm B\kern-.05em{\sc i\kern-.025em b}\kern-.08em
		T\kern-.1667em\lower.7ex\hbox{E}\kern-.125emX}}
\begin{document}
\title{Reinforcement Learning
with Soft Temporal Logic Constraints Using Limit-Deterministic Generalized B\"uchi Automaton

\thanks{$^{1}$Department of Mechanical Engineering, Lehigh University, Bethlehem, PA, 18015, USA. $^{2}$Department of Mechanical Engineering, University of Iowa Technology Institute, The University of Iowa,
Iowa City, IA, 52246, USA. $^{3}$Department of Automation, University of Science and Technology
of China, Hefei, Anhui, 230026, China.}
}

\author{Mingyu Cai$^{1}$, Shaoping Xiao$^{2}$, Zhijun Li$^{3}$, and Zhen Kan$^{3}$}
\maketitle

\begin{abstract}
This paper studies the control synthesis of motion
planning subject to uncertainties. The uncertainties are considered
in robot motions and environment properties, giving rise to the probabilistic
labeled Markov decision process (PL-MDP). A Model-Free Reinforcement
The learning (RL) method is developed to generate a finite-memory control policy
to satisfy high-level tasks expressed in linear temporal logic (LTL)
formulas. Due to uncertainties and potentially
conflicting tasks, this work focuses on infeasible LTL specifications, 
where a relaxed LTL constraint is developed to allow the agent to revise its motion plan and take violations of original tasks into account for partial satisfaction. And a novel automaton is developed to improve the density of accepting rewards and enable deterministic policies.
We proposed an RL framework with rigorous analysis that is guaranteed to achieve multiple objectives in decreasing order: 1) satisfying the acceptance
condition of relaxed product MDP and 2) reducing the violation cost over long-term behaviors. We provide simulation and experimental results to validate the performance.
\global\long\def\Dist{\operatorname{Dist}}%
\global\long\def\Inf{\operatorname{Inf}}%
\global\long\def\Sense{\operatorname{Sense}}%
\global\long\def\Eval{\operatorname{Eval}}%
\global\long\def\Info{\operatorname{Info}}%
\global\long\def\ResetRabin{\operatorname{ResetRabin}}%
\global\long\def\Post{\operatorname{Post}}%
\global\long\def\Acc{\operatorname{Acc}}%
\end{abstract}

	
\section{INTRODUCTION}

Formal logic is capable of describing complex high-level
tasks beyond traditional go-to-goal navigation for robotic systems.
As a formal language, linear temporal logic (LTL) has been increasingly
used in the motion planning of robotic systems \cite{Baier2008,Kloetzer2015,Kantaros2020}.
Since robotic systems are often subject to a variety of uncertainties
arising from the stochastic behaviors of the motion (e.g., an agent
may not exactly follow the control inputs due to potential sensing
noise or actuation failures) and uncertain environment properties
(e.g., there exist mobile obstacles or time-varying areas of interest),
Markov decision processes (MDPs) are often used to model the probabilistic
motion of robotic systems \cite{Thrun2005}. Based on probabilistic
model checking, control synthesis of MDP with LTL motion specifications
has been widely investigated (cf. \cite{Ding2011,Ding2014a,Ulusoy2014a,Guo2018,Lacerda2019,Vasile2020}).
In particular, the topic of partial satisfaction of high-level tasks in deterministic  and stochastic systems is investigated in \cite{Vasile2017,Cai2020receding,Lacerda2019,Cai2020optimal}.
Yet, new challenges arise when considering motion and environment
uncertainties. Hence, learning to find a satisfying policy is paramount
for the robot to operate in the presence of motion and environment
uncertainties.

Reinforcement learning (RL) is a sequential decision-making
process in which an agent continuously interacts with and learns from
the environment \cite{Watkins1992}. Model-based RL has been employed for motion planning with LTL specifications when full knowledge
of MDP is available \cite{Brazdil2014}. The work of \cite{Fu2014}
extends model-based RL to temporal logic constrained control of stochastic
systems with unknown parameters by model approximation. In \cite{Sadigh2014}
and \cite{Wang2015}, transition probabilities are learned to facilitate
the satisfaction of LTL specifications. 
However, these aforementioned
works have to reply on the accuracy of transition probabilities for learning.
On the other hand, model-free RL approaches with LTL-based rewards
generate desired policies by directly optimizing the Q-values~\cite{Aksaray2016,Li2019,Hahn2019,Bozkurt2019}.
However, these works are based on a key assumption that 
at least an accepting maximum end component (AMEC) exists in a standard product
MDP~\cite{Baier2008}, which may not be true in practice. For instance,
some areas of interest to be visited can be probabilistically prohibitive
to the agent in practice (e.g., potentially surrounded by water due
to heavy rain that the ground robot cannot traverse), resulting in
part of the user-specified tasks cannot be achieved and AMECs do not
exist in the product MDP. Although minimal revision of motion plans
in a potentially conflicting environment has been investigated in
the works of~\cite{Kim2012,Kim2012a,Tumova2013,Guo2015},
only deterministic transition systems are considered, and it is not
yet clear how to address the above-mentioned issues in stochastic
systems i.e., MDP.

\textbf{Related Works: } Compared with most RL-based approaches \cite{Sadigh2014, Hahn2019, Bozkurt2019, Cai2020maximum},
the reward functions are generally designed to enforce the convergence
to AMECs so that the acceptance condition can be satisfied. However,
these approaches are not applicable in this work since AMECs may
not even exist due to environmental uncertainties. Thus, the probability
of satisfying the pre-specified tasks may become zero. As an alternative
solution, the work of \cite{Gao2019} proposes a reduced variance
deep Q-learning method to approximate optimal state-action values. In addition,
 our previous work~\cite{Cai2020} develops a relaxed product MDP,
in which LTL formulas are converted into Deterministic Rabin Automaton
(DRA) without assuming the existence of AMECs. However, the authors
in \cite{Hahn2019} claim that it may fail to find desired policies
by converting LTL into DRA. Even though our recent work~\cite{cai2022learning} shows how to leverage planning algorithms to guide minimally-violating policies for infeasible LTL tasks. But multi-objective reinforcement learning without the assistance of planning is still under investigation.

In \cite{Hasanbeig2019a}, the LDGBA is
applied, and a synchronous frontier function for RL reward is designed
to synthesize control policies that mostly fulfill the given task
by maximizing the visits of the accepting sets in LDGBA.
However, it cannot record the visited or non-visited
accepting sets in each round. As an extension of~\cite{Cai2020},
this work can handle the above issues by developing an E-LDGBA to
record the non-visited accepting sets without adding extra computational
complexity. The proposed relaxed product MDP and the designed utility
function transforms the control synthesis problem into an expected utility
optimization problem, in which a satisfactory policy is guaranteed
to be found by leveraging RL to optimize the expected utility. Instead
of DRA, LDGBA is used to reduce the size of the automaton. It is well
known that the Rabin automaton, in the worst case, is doubly exponential
in the size of the LTL formula, while LDGBA for many LTL formulas
is an exponentially-sized automaton~\cite{Sickert2016}. Moreover,
the model-free RL-based approach is adopted and can synthesize optimal
policies on-the-fly without explicitly memorizing structures of relaxed
product MDP.

\textbf{Contributions: }
Motivated by these challenges, this work considers the transition probabilities of interactions between the environment and the mobile robot to be unknown. We study 
learning-based motion planning subject to uncertainties, where control objectives are defined as high-level LTL formulas to express complex tasks. The contributions of this work are multi-fold.
(i). From the aspect of automaton theory, we leverage an embedded limit
deterministic generalized B\"uchi automaton (E-LDGBA) that has several accepting sets to maintain a dense reward and allows applying deterministic policies to achieve high-level objectives.
(ii) Both motion and workspace uncertainties are considered to be unknown, leading to
potentially conflicting tasks (i.e., the pre-specified LTL tasks cannot
be fully satisfied).
We design a relaxed product MDP from the PL-MDP and the novel automaton so that the RL agent can revise its motion plan without strictly following the desired LTL constraints. 
(iii) An expected return
composed of violation rewards and accepting rewards is developed.
The designed violation function quantifies the differences between
the revised and the desired motion plans, while the accepting rewards
enhance the satisfaction of the acceptance condition of the relaxed
product MDP. 
(iv) Rigorous analysis is provided to show how to properly design parameters of accepting and violating rewards for multi-objective RL (MORL). Based on that, the RL agent can find policies that fulfill pre-specified tasks as much as possible.

\section{PRELIMINARIES}

\subsection{Probabilistic Labeled MDP\label{subsec:Labeled-MDP}}

\begin{defn}
\label{def:MDP}A probabilistic labeled MDP (PL-MDP) is a tuple $\mathcal{M}=(S,A,p_{S},(s_{0},l_{0}),\Pi,L,p_{L}, \varLambda )$,
where $S$ is a finite state space; $A$ is a finite action space;
$p_{S}:S\times A\times S\shortrightarrow\left[0,1\right]$ is the
transition probability function; $\Pi$ is a set of atomic propositions;
and $L:S\shortrightarrow2^{\Pi}$ is a labeling function. Let $\xi$
be an action function, which can be either deterministic such that
$\xi:S\rightarrow A$ maps a state $s\in S$ to an action in $A(s)$,
or randomized such that $\xi:S\times A\rightarrow\left[0,1\right]$
represents the probability of taking an action in $A(s)$
at $s$. The pair $(s_{0},l_{0})$ denotes an initial state
$s_{0}\in S$ with an initial label $l_{0}\in L(s_{0})$.
The function $p_{L}(s,l)$ denotes the probability of $l\in L(s)$
associated with $s\in S$ satisfying $\sum_{l\in L(s)}p_{L}(s,l)=1,\forall s\in S$.
The transition probability $p_{S}$ captures the motion uncertainties
of the agent while the labeling probability $p_{L}$ captures the
environment uncertainties.
\end{defn}

It is assumed that the agent can fully
observe its current state and the associated labels. 
The PL-MDP can be regarded as an advanced MDP, and we can use the MDP to denote it in the following sections. The MDP $\mathcal{M}$ evolves by taking actions $\xi_{i}$ at each
stage $i$, where
$i\in\mathbb{N}_{0}$ 
with $\mathbb{N}$ being the set of natural numbers.

\begin{defn}
The control policy $\boldsymbol{\mu}=\mu_{0}\mu_{1}\ldots$ is a
sequence of decision rules, which yields a path $\boldsymbol{s}=s_{0}s_{1}s_{2}\ldots$
over $\mathcal{M}$ with $p_{S}(s_{i},a_{i},s_{i+1})>0$
for all $i$. 
$\boldsymbol{\mu}$ can be either deterministic such that $\boldsymbol{\mu}:S\rightarrow A$
or stochastic such that $\boldsymbol{\mu}:S\times A\rightarrow\left[0,1\right]$.
The control policy $\boldsymbol{\mu}$ is memoryless if each $\mu_{i}$
only depends on its current state $s_{i}$. In contrast, $\boldsymbol{\mu}$
is called a finite-memory (i.e., history-dependent) policy if $\mu_{i}$ depends on its past
states. 
\end{defn}

This work shows how to apply the deterministic and memoryless policy that has more stable decision-making performance. Let $\boldsymbol{\mu}(s)$ denote the probability distribution of actions at state $s$, and $\boldsymbol{\mu}(s, a)$ represents the probability of generating action $a$ at state $s$ using the policy $\boldsymbol{\mu}$. Let $\varLambda:S\shortrightarrow\mathbb{R}$ denote a reward function
over $\mathcal{\mathcal{M}}$. Given a discount factor $\gamma\in(0,1)$,
the expected return under policy $\xi$ starting from $s\in S$ can
be defined as
\begin{equation}
U^{\xi}(s)=\mathbb{E}^{\xi}\left[\stackrel[i=0]{\infty}{\sum}\gamma^{i}\varLambda(s_{i})\left|s_{0}=s\right.\right]\label{eq:RL_return}
\end{equation}
The optimal policy $\xi^{*}$ is a policy that maximizes the expected
return for each state $s\in S$ as $\xi^{*}=\underset{\xi}{\arg\max}U^{\xi}(s).$

\begin{defn}
Given a PL-MDP $\mathcal{M}$ under policy $\boldsymbol{\pi}$, a  Markov
chain $MC_{\mathcal{M}}^{\boldsymbol{\mu}}$ of the PL-MDP $\mathcal{M}$ induced by a policy $\boldsymbol{\mu}$ is a tuple $(S,A,p^{\boldsymbol{\mu}}_{S},(s_{0},l_{0}), L, p_{L})$ where $p^{\boldsymbol{\mu}}_{S}(s, s')=p_{S}(s,a,s')$ with $\boldsymbol{\mu}(s,a)>0$ for all $s,s' \in S$.
\end{defn}

A sub-MDP $\mathcal{M}{}_{(S',A')}$ of $\mathcal{M}$ is a pair $(S', A')$ where $S'\subseteq S$ and $A'$ is a finite action space of $S'$ such that (i) $S'\neq\emptyset$, and $A'(s)\neq\emptyset , \forall s\in S'$; (ii) $\left\{ s' \mid  \forall s\in S' \text{ and }  \forall a\in A'(x), p^{\mathcal{P}}(s,a,s')>0\right\}\subseteq S'$. An induced graph of $\mathcal{M}{}_{(S',A')}$ is denoted as $\mathcal{G}_{(S',A')}$ that is a directed graph, where if $p_{S}(s,a,s')>0$ with $a\in A'(s)$, for any $s,s'\in S'$, there exists an edge between $s$ and $s'$ in $\mathcal{G}_{(S',A')}$. Note the evolution of a sub-MDP $\mathcal{M}{}_{(S',A')}$ is restricted by the action space $A'$.
A sub-MDP is a strongly connected component (SCC) if its induced graph is strongly connected such that for all pairs of nodes $s,s' \in S'$, there is a path from $s$ to $s'$. A bottom strongly connected component (BSCC) is an SCC
from which no state outside is reachable by applying the restricted action space. More details of the MDP treatments can be found in~\cite{Baier2008}.

\begin{defn}
\cite{Baier2008} A sub-MDP $\mathcal{M}{}_{(S',A')}$
is an end component (EC) of $\mathcal{M}$ if it's a BSCC. An EC $\mathcal{M}{}_{(S',A')}$ is called a maximal end
component (MEC) if there is no other EC $\mathcal{M}{}_{(S'',A'')}$
such that $S'\subseteq S''$ and $A'(s)\subseteq A''(s)$,
$\forall s\in S$.
\end{defn}

\subsection{LTL and Limit-Deterministic Generalized B\"uchi Automaton}

Linear temporal logic is a formal language to describe the high-level specifications of a system. An LTL formula is built on a set of atomic propositions, e.g., $a\in\AP$, standard Boolean
operators such as $\land$ (conjunction), $\lnot$ (negation), and
temporal operators $\mathcal{U}$ (until), $\varbigcirc$ (next),
$\diamondsuit$ (eventually), $\boxempty$ (always). The syntax of an LTL formula is defined inductively as
\begin{equation*}
        \phi   :=  \text{True} \mid a \mid \phi_1 \land \phi_2 \mid \lnot \phi \mid \next\phi \mid \phi_1 \UntilOp \phi_2\:, 
\end{equation*}
The semantics of an LTL formula are interpreted over words, which is an
infinite sequence $\boldsymbol{o}=o_{0}o_{1}\ldots$ where $o_{i}\in2^{\AP}$ for
all $i\geq0$, and $2^{\AP}$ represents the power set of $\AP$, which are defined as:
\begin{equation*}
\arraycolsep=1.4pt
\begin{array}{lcl}
\boldsymbol{o} \models \text{true}  \\
\boldsymbol{o} \models \alpha  & \Leftrightarrow & \alpha\in  L(\boldsymbol{o}[0])  \\
\boldsymbol{o} \models \phi_{1}\land\phi_{2} &  \Leftrightarrow & \boldsymbol{o} \models \phi_{1} \text{ and } \boldsymbol{o} \models \phi_{2}  \\
\boldsymbol{o} \models \lnot\phi  & \Leftrightarrow & \boldsymbol{o} \mid\neq\phi  \\
\boldsymbol{o} \models \next\phi  & \Leftrightarrow & \boldsymbol{o}[1:] \models\phi  \\
\boldsymbol{o} \models \phi_1 \UntilOp \phi_2  & \Leftrightarrow & \exists t \text{ s.t. }\boldsymbol{o}[t:]\models\phi_{2}, \forall t'\in [0,t),  \boldsymbol{o}[t':]\models\phi_{1}  \\
\end{array} 
\end{equation*}
Denote by $\boldsymbol{o}\models\varphi$ if the word $\boldsymbol{o}$
satisfies the LTL formula $\varphi$. More expressions can be achieved
by combing temporal and Boolean operators. Detailed descriptions of
the syntax and semantics of LTL can be found in \cite{Baier2008}.
Given an LTL formula that specifies the missions,
the satisfaction of the LTL formula can be evaluated by an LDGBA \cite{Sickert2016}.
\begin{defn}
	\label{def:GBA} A GBA is a tuple $\mathcal{A}=(Q,\Sigma,\delta,q_{0},F)$,
	where $Q$ is a finite set of states, $\Sigma=2^{\Pi}$ is a finite
	alphabet; $\delta\colon Q\times\Sigma\shortrightarrow2^{Q}$ is a
	transition function, $q_{0}\in Q$ is an initial state, and $F=\left\{ F_{1},F_{2},\ldots,F_{f}\right\} $
	is a set of acceptance conditions with $F_{i}\subseteq Q$, $\forall i\in\left\{ 1,\ldots f\right\} $. 
\end{defn}
Denote by $\boldsymbol{q}=q_{0}q_{1}\ldots$ a run of a GBA, where
$q_{i}\in Q$, $i=0,1,\ldots$. The run $\boldsymbol{q}$ is accepted
by the GBA, if it satisfies the generalized B\"uchi accepting sets,
i.e., $\inf(\boldsymbol{q})\cap F_{i}\neq\emptyset$, $\forall i\in\left\{ 1,\ldots f\right\} $,
where $\inf(\boldsymbol{q})$ denotes the set of states
that is visited infinitely often.

\begin{defn}
	\label{def:LDGBA} A GBA is a Limit-deterministic Generalized B\"uchi
	automaton (LDGBA) if $\delta\colon Q\times(\Sigma\cup\left\{ \epsilon\right\} )\shortrightarrow2^{Q}$
	is a transition function, and the states $Q$ can be partitioned into
	a deterministic set $Q_{D}$ and a non-deterministic set $Q_{N}$,
	i.e., $Q_{D}\cup Q_{N}=Q$ and $Q_{D}\cap Q_{N}=\emptyset$, where
	\begin{itemize}
		\item the state transitions in $Q_{D}$ are total and restricted within
		it, i.e., $\bigl|\delta(q,\alpha)\bigr|=1$ and $\delta(q,\alpha)\subseteq Q_{D}$
		for every state $q\in Q_{D}$ and $\alpha\in\Sigma$,
		\item the $\epsilon$-transitions  are only defined for state transitions from $Q_{N}$ to $Q_{D}$, and are not allowed in the deterministic set
		i.e., for any $q\in Q_{D}$, $\delta(q,\epsilon)=\emptyset,\forall\epsilon\in\left\{ \epsilon\right\}$,
		\item the accepting states are only in the deterministic set, i.e., $F_{i}\subseteq Q_{D}$
		for every $F_{i}\in F$.
	\end{itemize}
\end{defn}
In Def. \ref{def:LDGBA}, the $\epsilon$-transitions are only defined
for state transitions from $Q_{N}$ to $Q_{D}$ that do not consume
the input atomic proposition. Readers
are referred to \cite{Hahn2013} for algorithms with free implementations to convert an LTL formula to an LDGBA,.
Note that the state-based LDGBA
is used in this work for demonstration purposes. Transition-based
LDGBA can be constructed based on basic graph transformations. More
discussions about state-based and transition-based LDGBA in HOA format can be found in Owl \cite{Kretinsky2018}.

\begin{rem}
Unlike the widely used deterministic Rabin Automaton (DRA), LDGBA
has the Generalized B\"uchi Accepting condition that purely involves reachability problems. On the other hand, compared with Limit-Deterministic B\"uchi Automaton (LDBA) applied in~\cite{Bozkurt2019,Hahn2019}, LDGBA has more accepting sets shown in Fig.\ref{fig:example1} (a) to increase the density of rewards since the positive rewards are always assigned to the accepting states to enforce the acceptance condition. 
\end{rem}

\section{Problem Statement and Challenge}

The task specification to be performed by the agent
	is described by an LTL formula $\phi$ over $\Pi$. Given $\phi$,
$\mathcal{M},$ and $\boldsymbol{\xi}=\xi_{0}\xi_{1}\ldots$, the
induced infinite path is denoted by $\boldsymbol{s}_{\infty}^{\boldsymbol{\xi}}=s_{0}\ldots s_{i}s_{i+1}\ldots$
that satisfies $s_{i+1}\in\left\{ s\in S\bigl|p_{S}(s_{i},a_{i},s)>0\right\} $.
Let $L(\boldsymbol{s}_{\infty}^{\boldsymbol{\xi}})=l_{0}l_{1}\ldots$
be the sequence of labels associated with $\boldsymbol{s}_{\infty}^{\boldsymbol{\xi}}$
such that $l_{i}\in L(s_{i})$ and $p_{L}(s_{i},l_{i})>0$.
Denote by $L(\boldsymbol{s}_{\infty}^{\boldsymbol{\xi}})\models\phi$
if $\boldsymbol{s}_{\infty}^{\boldsymbol{\xi}}$ satisfies $\phi$.
The satisfaction probability under $\boldsymbol{\mu}$ from an initial
state $s_{0}$ can be computed as
\begin{equation}
\mathbb{\Pr}_{\mathcal{M}}^{\boldsymbol{\mu}}(\phi)=\mathbb{\Pr}_{\mathcal{M}}^{\boldsymbol{\mu}}(\boldsymbol{s}_{\infty}^{\boldsymbol{\mu}}\in\boldsymbol{S}_{\infty}^{\boldsymbol{\mu}} \left.\right|L(\boldsymbol{s}_{\infty}^{\boldsymbol{\mu}})\models\phi),\label{eq:SatisProb}
\end{equation}
where $\boldsymbol{S}_{\infty}^{\boldsymbol{\mu}}$ is
a set of all admissible paths under policy $\boldsymbol{\mu}$, and the computation of $\mathbb{\Pr}_{\mathcal{M}}^{\boldsymbol{\mu}}(\boldsymbol{s}_{\infty}^{\boldsymbol{\mu}})$ can be found in~\cite{Baier2008}. 

\begin{defn}
\label{def:feasibility}	
Given a PL-MDP $\mathcal{M}$, an LTL task
$\phi$ is fully feasible if and only if $\mathbb{\Pr}_{\mathcal{M}}^{\boldsymbol{\mu}}(\phi)>0$ s.t. there exists a path $\boldsymbol{s}_{\infty}^{\boldsymbol{\mu}}$ over the infinite horizons under the policy $\boldsymbol{\mu}$ satisfying $\phi$.
\end{defn}	

\begin{figure}[!t]\centering
	\subfloat[]{{\includegraphics[scale=0.30]{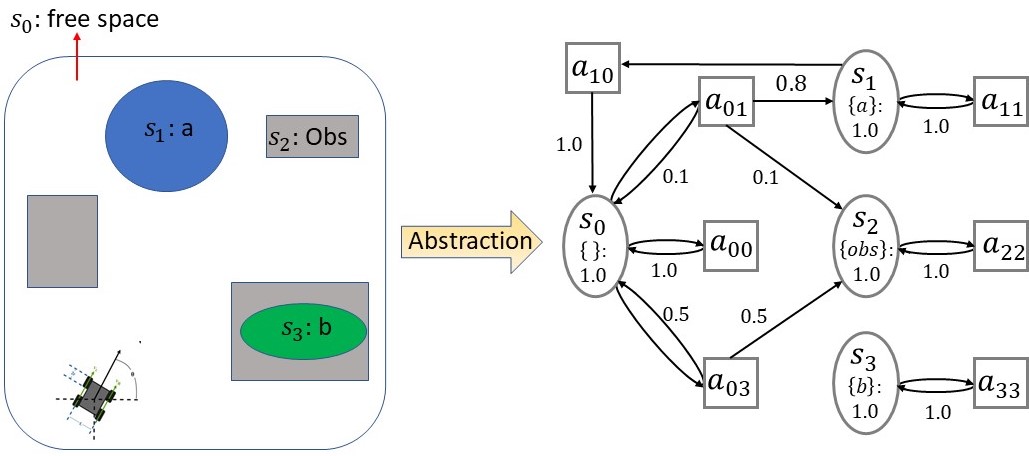} }}
	$\qquad$
	\subfloat[]{{
	\includegraphics[scale=0.40]{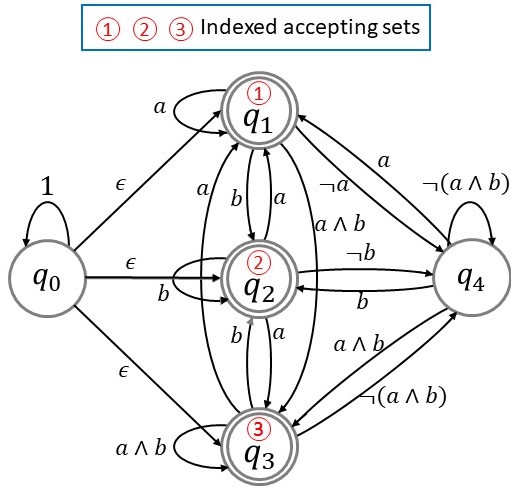}
		}}
	\caption{\label{fig:example1}  A running example of an infeasible LTL task with respect to the PL-MDP. (a) The abstraction process from a robotic system into a PL-MDP $\mathcal{M}$, where the region labeled with $b$ is surrounded by obstacles and can not be accessed. As a result, $\phi$ is infeasible in this case. (b) The LDGBA of an LTL formula $\phi=\oblong\lozenge\mathtt{a}\land\oblong\lozenge\mathtt{b}$.}
\end{figure}

Based on Def.~\ref{def:feasibility}, an infeasible case means there does not exist any policy $\mu$ to satisfy the task i.e., $\mathbb{\Pr}_{\mathcal{M}}^{\boldsymbol{\mu}}(\phi)=0$. 

\begin{example}
To illustrate an infeasible case, the environment, its abstracted PL-MDP, and the LDGBA of an LTL formula $\phi=\oblong\lozenge\mathtt{a}\land\oblong\lozenge\mathtt{b}$
are shown in Fig.~\ref{fig:example1} (a) and (b), respectively.
Since obstacles surround region $s_{3}$, the given LTL task is infeasible in this case.
\end{example}

We define the expected discount violation cost of task satisfaction as follows, considering feasible and infeasible tasks.

\begin{defn}
\label{def:infeasibility}	
Given a PL-MDP $\mathcal{M}$ and an LTL task $\phi$, the discount expected violation under the policy $\boldsymbol{\mu}$ is defined as
\begin{equation}
J_{V}(\mathcal{M}^{\boldsymbol{\mu}},\phi)=\mathbb{E}_{\mathcal{M}}^{\boldsymbol{\mu}}\left[\stackrel[i=0]{\infty}{\sum}\gamma^{i}c_{V}(s_{i}, a_{i},s_{i+1},\phi)\right],\label{eq:AVPS1}
\end{equation}
where $c_{V}(s,a,s',\phi)$ is defined as the violation cost of a transition $(s,a,s')$ with respect to $\phi$, and $a_{i}$ is the action generated based on the policy $\boldsymbol{\mu}(s_{i})$.
\end{defn}
Most existing results (cf.
\cite{Sadigh2014,Hahn2019,Bozkurt2019}) assume the existence
of at least one policy $\mu$ s.t. $\mathbb{\Pr}_{\mathcal{M}}^{\boldsymbol{\mu}}(\phi)>0$, which may not be true in practice
if the task is only partially feasible. In addition, previous work~\cite{Cai2020optimal} addresses the infeasible cases via an optimization approach that assumes the transitions probabilities of $\mathcal{M}$ are known. We remove this assumption and apply the RL approach to learn the desired policies in this work. We consider the following problem to account for challenges of both infeasible cases and unknown transitions.

\begin{problem}
\label{Prob1} 
Given an LTL-specified
task $\phi$ and a PL-MDP with unknown transition
probabilities (i.e., motion uncertainties) and unknown probabilistic
label functions (i.e., workspace uncertainties), the objective is
to find a policy (1) $\mathbb{\Pr}_{\mathcal{M}}^{\boldsymbol{\mu}}(\phi)>0$ if it exists; (2) if $\forall\boldsymbol{\mu}, \mathbb{\Pr}_{\mathcal{M}}^{\boldsymbol{\mu}}(\phi)=0$, find the policy $\boldsymbol{\mu}$ that 
mostly fulfills the desired task $\phi$ (i.e.,
infeasible constraints) over an infinite horizon by minimizing $J_{V}(\mathcal{M}^{\boldsymbol{\mu}},\phi)$.
\end{problem}
	
\section{Automaton Analysis}

To solve Problem \ref{Prob1}, Section \ref{Embedded-LDGBA}
first presents how the LDGBA in Def. \ref{def:LDGBA} can be extended
to E-LDGBA to keep tracking the non-visited accepting sets. Section~\ref{subsec:Pro_MDP} shows the traditional approach and the corresponding challenges for infeasible tasks. Section~\ref{subsec:Relax_PMDP} presents the construction of a relaxed product
MDP to handle soft LTL constraints. The benefits of incorporating
E-LDGBA with relaxed product MDP are discussed in Section \ref{subsec:Property_relax}.
	
\subsection{E-LDGBA\label{Embedded-LDGBA}}

In order to find the desired policy in PL-MDP $\mathcal{M}$ to satisfy the user-specified LTL formula $\phi$, one can construct the standard product MDP between $\mathcal{M}$ and the LDGBA of $\phi$ as described in \cite{Baier2008}. Then, the problem becomes finding the policy that satisfies the accepting condition of the standard product MDP. However, directly adopting LDGBA may fail to satisfy the LTL specifications when applying deterministic policies. More details can be found in our conference paper \cite{Cai2020maximum}. To overcome the issues, the E-LDGBA is introduced as follows.

Given an LDGBA $\mathcal{A}=(Q,\Sigma,\delta\cup\left\{ \epsilon\right\},q_{0},F)$,
a tracking-frontier set $T$ is
designed to keep track of non-visited accepting sets. Particularly,
$T$ is initialized as $F$, which is then updated based on 
\begin{equation}
	f_{V}(q,T)=\left\{ \begin{array}{cc}
		T\setminus F_{j}, & \text{if }q\in F_{j}\text{ and }F_{j}\in T,\\
		F\setminus F_{j}, & \text{if }\ensuremath{q\in F_{j}\text{ and }T=\emptyset},\\
		T, & \text{otherwise. }
	\end{array}\right.\label{eq:Trk-fontier}
\end{equation}
Once an accepting set $F_{j}$ is visited, it will be removed from
$T$. If $T$ becomes empty, it will be reset as $F\setminus F_{j}$. Since the acceptance
condition of LDGBA requires infinitely visiting all accepting sets,
we call it one round if all accepting sets have been visited (i.e.,
a round ends if $T$ becomes empty). If a state $q$ belongs to multiple
sets of $T$, all of these sets should be removed from $T$.

\begin{defn}[Embedded LDGBA]
\label{def:E-LDGBA} Given an LDGBA $\mathcal{A}=(Q,\Sigma,\delta\cup\left\{ \epsilon\right\},q_{0},F)$,
its corresponding E-LDGBA is denoted by $\mathcal{\overline{A}}=(\overline{Q},\Sigma,\overline{\delta}\cup\left\{ \epsilon\right\},\overline{q_{0}},\overline{F},f_{V},T)$
where
\begin{compactitem}[$\bullet$]
\item $T$ is initially set as $T=F$;
\item $\overline{Q}=Q\times2^{F}$ is the set of augmented states e.g., $\overline{q}=(q,T)$; The finite alphabet $\Sigma$ is the same as the one in the LDGBA;
\item The transition $\overline{\delta}\colon \overline{Q}\times(\Sigma\cup\left\{ \epsilon\right\} )\shortrightarrow2^{\overline{Q}}$
is defined as $\overline{q'}=\overline{\delta}(\overline{q},\overline{\sigma})$ with $\overline{\sigma}\in(\Sigma\cup\left\{ \epsilon\right\})$, e.g., $\overline{q}=(q,T)$ and $\overline{q'}=(q',T)$, and it satisfies
two conditions: 1) $q'=\delta(q,\overline{\sigma})$, and 2) $T$ is synchronously updated as $T=f_{V}(q',T)$ after transition $\overline{q'}=\overline{\delta}(\overline{q},\alpha)$;
\item $\overline{F}=\left\{\overline{F_{1}},\overline{F_{2}}\ldots \overline{F_{f}}\right\}$ is a set of accepting states,
where $\overline{F_{j}}=\left\{ (q, T)\in \overline{Q}\bigl|q\in F_{j}\land F_{j} \subseteq T\right\} $,
$j=1,\ldots f$.    
\end{compactitem}
\end{defn}

In Definition \ref{def:E-LDGBA}, we abuse the tuple structure since the frontier set $T$ is synchronously updated after each transition. \textcolor{black}{The state-space is augmented with the tracking-frontier set $T$ that can be practically represented via one-hot encoding based on the indices of the accepting set.} The accepting state is determined based on the current automaton state and the frontier set $T$. 
 Such property
is the innovation of E-LDGBA, which encourages all accepting sets to be
visited in each round. 
Alg.~\ref{Alg:LDGBA} demonstrates the procedure
of obtaining a valid run $\boldsymbol{q}_{\overline{\mathcal{A}}}$ over E-LDBGA $\mathcal{\overline{A}}_{\phi}$. Given an input alphabet $\alpha$ at each time step, lines $6-10$ show how to generate the next state of $\mathcal{\overline{A}}_{\phi}$ and update the tracking frontier set $T$ synchronously. Also, lines $6-8$ indicate the difference and relationship between $\mathcal{\overline{A}}_{\phi}$ and its corresponding LDGBA $\mathcal{A}_{\phi}$.

Given $\mathcal{\overline{A}}_{\phi}$ and $\mathcal{A}_{\phi}$ for the same LTL formula, the E-LDGBA $\mathcal{\overline{A}}_{\phi}$ keeps track of unvisited accepting sets of  $\mathcal{A}_{\phi}$ by incorporating $f_{V}$ and $T$, which can distinguish and enforce the procedure of acceptance satisfaction related to different accepting sets. $T$ will be reset when all the accepting sets of $\mathcal{A}_{\phi}$ have been visited. Let $\mathcal{L}(\mathcal{A}_{\phi})\subseteq \Sigma^{\omega}$ and $\mathcal{L}(\mathcal{\overline{A}}_{\phi})\subseteq \Sigma^{\omega}$ be the accepted language of $\mathcal{A}_{\phi}$ and $\mathcal{\overline{A}}_{\phi}$, respectively, with the same alphabet $\Sigma$. Based on \cite{Baier2008}, $\mathcal{L}(\mathcal{A}_{\phi})\subseteq \Sigma^{\omega}$ is the set of all infinite words accepted by $\mathcal{A}_{\phi}$ that satisfies LTL formula $\phi$.

\begin{algorithm}
	\caption{\label{Alg:LDGBA}Synthesizing a valid run of E-LDGBA}
	
	\singlespacing
	
	\begin{algorithmic}[1]
		
		\Procedure {Input: } {LDGBA $\mathcal{A}$, $f_{V},T$ and length
			$L$}
		
		{Output: } { A valid run $\overline{\boldsymbol{q}}_{\overline{A}}$ with
			length $L$ in $\mathcal{\overline{A}}_{\phi}$ }
		\State set $T=F$ and $count=1$
		
		\State set $\overline{q}_{cur}=(q_{0},T)$ and $\overline{\boldsymbol{q}}_{\overline{A}}=(q_{cur})$
		
		\State set $q_{cur}=q$ 
		
		\While { $count\leq L$ }
		
		\State given an input $\alpha$, $q_{next}=\delta(q_{cur},\alpha)$
		
		\State obtain $\overline{q}_{cur}\shortleftarrow (q_{next},T)$
		
		\State check if $\overline{q}_{cur}$ is an accepting state
		
	    \State add state $\overline{q}_{cur}$ to $\overline{\boldsymbol{q}}_{\overline{A}}$
		
		\State update $T=f_{V}(q_{next},T)$
		
		\State $count++$ and $q_{cur}\shortleftarrow q_{next}$
		
		\EndWhile
		
		\EndProcedure
		
	\end{algorithmic}
\end{algorithm}

\begin{lem}
\label{lem:language}
For any LTL formula $\phi$, we can construct LDGBA $\mathcal{A}_{\phi}=(Q,\Sigma,\delta,q_{0},F)$ and E-LDGBA $\mathcal{\overline{A}}_{\phi}=(\overline{Q},\Sigma,\overline{\delta},\overline{q_{0}},\overline{F},f_{V},T)$. Then it holds that
\begin{equation}
\mathcal{L}(\mathcal{\overline{A}}_{\phi})=\mathcal{L}(\mathcal{A}_{\phi}).
\end{equation}
\end{lem}

\begin{proof}
Details of the proof can be found in our work \cite{Cai2020maximum} that has shown $\mathcal{L}(\mathcal{\overline{A}}_{\phi})\supseteq\mathcal{L}(\mathcal{A}_{\phi})$ and $\mathcal{L}(\mathcal{\overline{A}}_{\phi})\subseteq\mathcal{L}(\mathcal{A}_{\phi})$. It indicates that E-LDGBA accept the same language as LDGBA, and can be applied to verify the LTL satisfaction.
\end{proof}

\subsection{Product MDP and Challenges\label{subsec:Pro_MDP}}

To find a policy $\boldsymbol{\mu}$ satisfying $\phi$ i.e., $\mathbb{\Pr}_{\mathcal{M}}^{\boldsymbol{\mu}}(\phi)>0$, one can construct the standard product MDP as follows.

\begin{defn}
\label{def:P-MDP} Given the labeled MDP $\mathcal{M}$ and the E-LDGBA
$\overline{\mathcal{A}}_{\phi}$ corresponding to $\phi$, the product MDP is
defined as $\overline{\mathcal{P}}=(Y,U^{\overline{\mathcal{P}}},p^{\overline{\mathcal{P}}},x_{0},F^{\overline{\mathcal{P}}})$,
where 
\begin{compactitem}[$\bullet$]
\item $Y=S\times2^{\Pi}\times \overline{Q}$ is the set of labeled states, i.e.,
$y=(s,l,\overline{q})\in Y$ with $l\in L(s)$ satisfying
$p_{L}(s,l)>0$; 
\item  $x_{0}=(s_{0},l_{0},\overline{q}_{0})$ is the initial
state;
\item $U^{\overline{\mathcal{P}}}=A\cup\left\{ \epsilon\right\} $
is the set of actions, where the $\epsilon$-actions
are only allowed for transitions of E-LDGBA components from $\overline{Q}_{N}$
to $\overline{Q}_{D}$;
\item $\overline{F}^{\overline{\mathcal{P}}}=\left\{ \overline{F}_{1}^{\overline{\mathcal{P}}},\overline{F}_{2}^{\overline{\mathcal{P}}}\ldots \overline{F}_{f}^{\overline{\mathcal{P}}}\right\}$ is the set of accepting states.
where $\overline{F}_{j}^{\overline{\mathcal{P}}}=\left\{ (s,l,\overline{q})\in X\bigl|\overline{q}\in \overline{F}_{j}\right\} $,
$j=1,\ldots f$;
\item  $p^{\overline{\mathcal{P}}}:Y\times U^{\overline{\mathcal{P}}}\times Y\shortrightarrow\left[0,1\right]$
is transition probability defined as: 1) $p^{\overline{\mathcal{P}}}(y,u^{\overline{\mathcal{P}}},y')=p_{L}(s',l')\cdotp p_{S}(s,a,s^{\prime})$
if $\delta(\overline{q},l)=\overline{q}^{\prime}$ and $u^{\overline{\mathcal{P}}}=a\in A(s)$;
2) $p^{\overline{\mathcal{P}}}(y,u^{\overline{\mathcal{P}}},y')=1$ if $\ensuremath{u^{\overline{\mathcal{P}}}\in\left\{ \epsilon\right\} }$,
$\overline{q}'\in\delta(\overline{q},\epsilon)$, and $(s',l')=(s,l)$;
and 3) $p^{\overline{\mathcal{P}}}(y,u^{\overline{\mathcal{P}}},y')=0$ otherwise.
\end{compactitem}
\end{defn}

The product MDP  $\overline{\mathcal{P}}$ captures the intersections between
all feasible paths over $\mathcal{M}$ and all words accepted to $\mathcal{A}_{\phi}$,
facilitating the identification of admissible agent motions that satisfy
the task $\phi$. Let $\boldsymbol{\pi}$ denote a policy over $\overline{\mathcal{P}}$
and denote by $\boldsymbol{y}_{\infty}^{\boldsymbol{\pi}}=y_{0}\ldots y_{i}y_{i+1}\ldots$
the infinite path generated by $\boldsymbol{\pi}$. A path $\boldsymbol{y}_{\infty}^{\boldsymbol{\pi}}$
is accepted if $\inf(\boldsymbol{y}_{\infty}^{\boldsymbol{\pi}})\cap \overline{F}_{i}^{\overline{\mathcal{P}}}\neq\emptyset$
, $\forall i\in\left\{ 1,\ldots f\right\} $. If $\boldsymbol{y}_{\infty}^{\pi}$
is an accepting run, there exists a policy $\boldsymbol{\xi}$ in
$\mathcal{M}$ that satisfies $\phi$.

Consider a sub-product MDP $\overline{\mathcal{P}}'_{(Y',U')}$, where
$Y'\subseteq X$ and $U'\subseteq U^{\overline{\mathcal{P}}}$. If $\overline{\mathcal{P}}'_{(Y',U')}$
is a MEC of $\overline{\mathcal{P}}$ and $Y'\cap \overline{F}_{i}^{\overline{\mathcal{P}}}\neq\emptyset$,
$\forall i\in\left\{ 1,\ldots f\right\} $, then $\overline{\mathcal{P}}'_{(Y',U')}$
is called an accepting maximum end component (AMEC) of $\overline{\mathcal{P}}$.
Once a path enters an AMEC, the subsequent path will stay within it
by taking restricted actions from $U'$. There exist
policies such that any state $x\in X'$ can be visited infinitely
often. As a result, satisfying the task $\phi$ is equivalent to reaching an AMEC. Moreover, a MEC that does not contain any accepting sets is called a rejecting accepting component (RMEC). 
A MEC with only partial accepting sets contained is called a neutral maximum end component (NMEC). 

\subsection{Relaxed Product MDP\label{subsec:Relax_PMDP}}

For the product MDP \textcolor{black}{$\overline{\mathcal{P}}$ or $\mathcal{\overline{\mathcal{P}}}$}
introduced above, the satisfaction of $\phi$ is based on the assumption
that at least one AMEC exists in the product MDP, i.e., at least one policy satisfies the given LTL formula with respect to the PL-MDP.
Otherwise, the task is infeasible with respect to the PL-MDP. We treat the LTL task as soft constraints to address the cases where tasks can be potentially infeasible. The relaxed product MDP is designed to allow the agent
to revise its motion plan without strictly following the desired LTL
constraints.

\begin{defn}
\label{def:relaxed-product} \sloppy The relaxed product MDP is constructed
from $\mathcal{\mathcal{\overline{\mathcal{P}}}}$ as a tuple $\overline{\mathcal{R}}=\mathcal{M}\otimes\mathcal{\overline{A}}_{\phi}=(X,U^{\overline{\mathcal{R}}},p^{\overline{\mathcal{R}}},x_{0},F^{\overline{\mathcal{R}}},c_{V}^{\overline{\mathcal{R}}},f_{V},T)$
, where

\begin{compactitem}[$\bullet$]
\item $X$, $x_{0}$, and $F^{\overline{\mathcal{R}}}=F^{\overline{\mathcal{P}}}$ are the same as in $\overline{\mathcal{P}}$;

\item $U^{\overline{\mathcal{R}}}$ is the set of extended actions that jointly consider the actions of $\mathcal{M}$ and
the input atomic proposition of $\ensuremath{\mathcal{\mathcal{A}_{\phi}}}$. Specifically, given a state $x=(s,l,q)\in X$,
the available actions are $u^{\overline{\mathcal{R}}}(x)=\left\{ (a,\iota)\bigl|a\in A(s),\iota\in(2^{\AP}\cup\left\{\epsilon\right\})\right\} $.
Given an action $u^{\overline{\mathcal{R}}}=(a,\iota)\in u^{\overline{\mathcal{R}}}(x)$,
the projections of $u^{\overline{\mathcal{R}}}$ to $A(s)$ in $\mathcal{M}$
and to $2^{\AP}\cup\left\{ \epsilon\right\} $ in $\mathcal{A}_{\phi}$
are denoted by $u\bigr|_{\mathcal{M}}^{\overline{\mathcal{R}}}$ and
$u\bigr|_{\ensuremath{\mathcal{A}}}^{\overline{\mathcal{R}}}$, respectively;

\text{ }

\item $p^{\overline{\mathcal{R}}}:X\times u^{\overline{\mathcal{R}}}\times X\shortrightarrow\left[0,1\right]$
	is the transition function. The
transition probability $p^{\overline{\mathcal{R}}}$ from a state $x=(s,l,\overline{q})$
to a state $x'=(s',l',\overline{q}')$ is defined as: 1) $p^{\overline{\mathcal{R}}}(x,u^{\overline{\mathcal{R}}},x')=p_{L}(s',l')\cdotp p_{S}(s,a,s^{\prime})$
with $a=u\bigr|_{\mathcal{M}}^{\mathcal{\overline{\mathcal{R}}}}$, if $\overline{q}$ can
be transitioned to $\overline{q}^{\prime}$ and $\overline{q}'=\overline{\delta}(\overline{q},u\bigr|_{\ensuremath{\overline{\mathcal{A}}}}^{\overline{\mathcal{R}}})$
s.t $u\bigr|_{\ensuremath{\overline{\mathcal{A}}}}^{\overline{\mathcal{R}}}\neq\epsilon$;
2) $p^{\overline{\mathcal{R}}}(x,u^{\overline{\mathcal{R}}},x')=1$, if $u\bigr|_{\ensuremath{\overline{\mathcal{A}}}}^{\overline{\mathcal{R}}}=\epsilon$,
$\overline{q}'\in\overline{\delta}(\overline{q},\epsilon)$, and $(s',l')=(s,l)$;
3) $p^{\overline{\mathcal{R}}}(x,u^{\overline{\mathcal{R}}},x')=0$ otherwise.
Under an action $u^{\overline{\mathcal{R}}}\in U^{\overline{\mathcal{R}}}(x)$,
it holds that $\sum_{x'\in X}p^{\overline{\mathcal{R}}}(x,u^{\overline{\mathcal{R}}},x')=1$;

\item $c_{V}^{\overline{\mathcal{R}}}:X\times u^{\overline{\mathcal{R}}}\times X\shortrightarrow\mathbb{R}$
is the violation cost. To define the violation cost $c_{V}^{\overline{\mathcal{R}}}$, consider $\Pi=\left\{ \iota_{1},\iota_{2}\ldots\iota_{M}\right\} $
and an evaluation function $\Eval\colon2^{\Pi}\shortrightarrow\left\{ 0,1\right\} ^{M}$,
defined as $\Eval(l)=\left[v_{i}\right]^{M}$ with $v_{i}=1$
if $\iota_{i}\in l$ and $v_{i}=0$ otherwise, where $i=1,2,\ldots,M$
and $l\in2^{\Pi}$. To quantify the difference between two elements
in $2^{\Pi}$, we introduce $\rho(l,l')=\left\Vert v-v^{\prime}\right\Vert _{1}=\sum_{i=1}^{M}\left|v_{i}-v_{i}^{\prime}\right|,$where
$v=\Eval(l)$, $v^{\prime}=\Eval(l^{\prime})$,
$l,l^{\prime}\in2^{\Pi}$, and $\left\Vert \cdot\right\Vert _{1}$
is the $\ell_{1}$ norm. The distance from $l\in2^{\Pi}$ to a set
$\mathcal{X}\subseteq2^{\Pi}$ is then defined as $\Dist(l,\ensuremath{\mathcal{X}})=0$
if $l\in\ensuremath{\mathcal{X}}$, and $\Dist(l,\ensuremath{\mathcal{X}})=\underset{l'\in\ensuremath{\mathcal{X}}}{\min}\rho(l,l')$
if $l\notin\ensuremath{\mathcal{X}}$. 
The violation cost of the transition from $x=(s,l,q)$
to $x'=(s',l',q')$ under an action $u^{\overline{\mathcal{R}}}$
is defined as
\[
c_{V}^{\overline{\mathcal{R}}}(x,u^{\overline{\mathcal{R}}},x')=\left\{ \begin{array}{cc}
p_{L}(s',l')\cdotp w_{V}(x,x') & \text{ if \ensuremath{u\bigr|_{\ensuremath{\mathcal{A}}}^{\overline{\mathcal{R}}}\neq\epsilon}, }\\
0 & \text{otherwise},
\end{array}\right.
\]
where $w_{V}(x,x')=\Dist(L(s),\ensuremath{\mathcal{X}}(q,q^{\prime}))$
with $\ensuremath{\mathcal{X}}(q,q')=\left\{ l\in2^{\boldsymbol{\pi}}\left|q\overset{l}{\shortrightarrow}q'\right.\right\} $
being the set of input alphabets that enables the transition from
$q$ to $q^{\prime}$. Borrowed from \cite{Guo2015}, the function
$\Dist(L(s),\ensuremath{\mathcal{X}}(q,q^{\prime}))$
measures the distance from $L(s)$ to the set $\ensuremath{\mathcal{X}}(q,q^{\prime})$.
\end{compactitem}
\end{defn}

\begin{algorithm}
	\caption{\label{Alg1}  Policy Execution over $\overline{\mathcal{R}}=\mathcal{M}\otimes\mathcal{\overline{A}}_{\phi}$}
	
	\singlespacing
	
	\begin{algorithmic}[1]
		
		\Procedure {Input: } {policy $\pi$, $\mathcal{M}$, $\mathcal{\overline{A}}_{\phi}$,
			$N$ }
		
		{Output: } {a valid run $\boldsymbol{x}_{N}$
			with length of $N$ under policy $\pi$ over $\overline{\mathcal{R}}$}
		
		\State set $t=0$ and $x_{t}=x_{0}$
		
		\State set $\boldsymbol{x}_{N}=(x_{t})$
		
		\While { $t<N$ }
		
		\State obtain $u_{t}^{\overline{\mathcal{R}}}$ based on $\pi(x_{t})$
		
		\State execute $u_{t}\bigr|_{\mathcal{M}}^{\mathcal{\overline{\mathcal{R}}}}$ for
    	$\mathcal{M}$ and obtain $s_{t+1}$

		\State observe label $l_{t+1}$ according to $L(s_{t})$
    	
		\State obtain $q_{t+1}=\delta(q_{t},u\bigr|_{\ensuremath{\overline{\mathcal{A}}}}^{\overline{\mathcal{R}}})$ s.t. $\overline{q}_{t+1}=( q_{t+1}, T)$
		
		\State obtain $x_{t+1}=(s_{t+1}, l_{t+1} q_{t+1}, T)$
		
		\State check if the $x_{t+1}$ is an accepting state
		
		\State compute $c_{V}^{\overline{\mathcal{R}}}(x,u^{\overline{\mathcal{R}}},x')$
		based on Def. \ref{def:relaxed-product}
	
		\State add $x_{t+1}$ to $\boldsymbol{x}_{N}$
		
		\State $x_{t}=x_{t+1}$ and $t++$
		
		\State update $T\shortleftarrow f_{V}(q',T)$ for $\mathcal{\overline{A}}_{\phi}$
		
		\EndWhile
		
		\EndProcedure
		
	\end{algorithmic}
\end{algorithm}

Alg.~\ref{Alg1} illustrates executing a given policy and generating a valid run over the relaxed product MDP $\overline{\mathcal{R}}$ on-the-fly. "On-the-fly" means the procedure does not need to construct $\overline{\mathcal{R}}$ completely. In contrast, it tracks the states based on the evolution in Def.~\ref{def:relaxed-product}. In line $8$ of Alg.~\ref{Alg1}, the next automaton state of $\mathcal{\overline{A}}_{\phi}$ is reached according to lines $6-7$ of Alg.~\ref{Alg:LDGBA} by assigning the input alphabet as $u\bigr|_{\ensuremath{\overline{\mathcal{A}}}}^{\overline{\mathcal{R}}}$. In lines 5-9, the next state is jointly determined by the transitions of PL-MDP and E-LDGBA (Alg.~\ref{Alg:LDGBA}). Line $14$ of Alg.~\ref{Alg1} also updates the tracking frontier set synchronously as Alg.~\ref{Alg:LDGBA}, and line $11$ computes its violation cost after generating the valid transition.

The weighted violation function $w_{V}(x,x')$
quantifies how much the transition from $x$ to $x'$ in a product
MDP violates the constraints imposed by $\phi$.
The transition probability $p^{\overline{\mathcal{R}}}$ jointly considers the
probability of an event $p_{L}(s',l')$ and the transition
$p_{S}$. Then the measurement of violation in~\eqref{eq:AVPS1} can be reformulated over relaxed product MDP as the expected return of violation cost:

\begin{defn}
Given a relaxed product MDP $\overline{\mathcal{R}}$ generated from a PL-MDP $\mathcal{M}$ and an E-LDBA $\overline{A}_{\phi}$, the measurement of expected violation in~\eqref{eq:AVPS1} can be reformulated over relaxed product MDP under policy $\boldsymbol{\pi}$ as:
\begin{equation}
J_{V}(\overline{\mathcal{R}}^{\boldsymbol{\pi}})=\mathbb{E}_{\overline{\mathcal{R}}}^{\boldsymbol{\pi}}\left[\stackrel[i=0]{\infty}{\sum}\gamma^{i}c_{V}^{\overline{\mathcal{R}}}(x_{i},u_{i}^{\overline{\mathcal{R}}},x_{i+1}).\right]\label{eq:violation}
\end{equation}
\end{defn}

Finding a policy to minimize $J_{V}(\overline{\mathcal{R}}^{\boldsymbol{\pi}})$
will enforce the planned path towards more fulfilling the LTL task
$\phi$ by penalizing $w_{V}$. A run $r_{\overline{\mathcal{P}}}^{\pi}$ induced
by a policy $\pi$ that satisfies the accepting condition of the relaxed
product MDP $\overline{\mathcal{R}}$ completes the corresponding LTL mission
$\phi$ exactly if and only if the violation return~\eqref{eq:violation}
is equal to zero.

\subsection{Properties of Relaxed Product MDP\label{subsec:Property_relax}}

This section discusses the properties of the relaxed product MDP and standard product MDP.
Applying the same process in Def.~\ref{def:P-MDP}, we can construct the product MDP between LDGBA and PL-MDP denoted as $\mathcal{P}=\mathcal{M}\otimes\mathcal{A}_{\phi}$. The detailed procedure can be found in~\cite{Baier2008}.
The relaxed product MDP between MDP and LDGBA can also be  constructed as the procedure Def.~\ref{def:relaxed-product} by replacing E-LDGBA with LDGBA denoted as $\mathcal{R}=\mathcal{M}\otimes\mathcal{A}_{\phi}$. Note, the product and relaxed product are the procedures of constructing the interaction between the automaton structure and the MDP model, which can be applied to any automata. 

Similarly, $\mathcal{R}$ and $\mathcal{P}$ for the same $\mathcal{A}_{\phi}$ and PL-MDP $\mathcal{M}$ share the same states. Hence, we can regard $\mathcal{R}$ and $\mathcal{P}$ as two separate directed graphs with the the same nodes. 
From the graph aspect, the MEC can be regarded as a bottom strongly connected component (BSCC). Similarly, let ABSCC denote the BSCC that intersects with all accepting sets in $\mathcal{R}$ or $\mathcal{P}$. Then, we have the following conclusion.

\begin{thm}
	\label{Thm_Properties}Given an MDP $\mathcal{M}$ and an
	LDGBA $\mathcal{A}_{\phi}$ of $\phi$, the relaxed product
	MDP $\mathcal{R}=\mathcal{M}\otimes\mathcal{A}_{\phi}$ and corresponding standard product MDP $\mathcal{P}$
	have the following properties:
	\begin{enumerate}
		\item the directed graph of standard product MDP $\mathcal{P}$ is a sub-graph of
		 the directed graph of $\mathcal{R}$,
		\item there always exists at least one AMEC in $\mathcal{R}$,
		\item if the LTL formula $\phi$ is feasible over $\mathcal{M}$, any direct graph of AMEC of $\mathcal{P}$ is the sub-graph of a direct graph of AMEC of $\mathcal{R}$.
	\end{enumerate}
\end{thm}
\begin{proof}
The proof is similar as our previous work \cite{Cai2020optimal} by replacing the LDBA with LDGBA.
\end{proof}

Since the E-LDGBA is an extension of LDGBA to enable the recording ability regarding accepting sets, which accept the same LTL formulas as in Lemma~\ref{lem:language}. We can also have the same conclusion for $\overline{\mathcal{R}}$ and $\overline{\mathcal{P}}$.

\begin{lem}
\label{lem:Properties}Given an MDP $\mathcal{M}$ and an
E-LDGBA $\overline{\mathcal{A}}_{\phi}$ of $\phi$, the relaxed product
MDP $\overline{\mathcal{R}}=\mathcal{M}\otimes\overline{\mathcal{A}}_{\phi}$ and its corresponding $\overline{\mathcal{P}}$ have the following properties
\begin{enumerate}
\item the directed graph of standard product $\overline{\mathcal{P}}$ is a sub-graph of
 the directed graph of $\overline{\mathcal{R}}$,
\item there always exists at least one AMEC in $\overline{\mathcal{R}}$,
\item if the LTL formula $\phi$ is feasible over $\mathcal{M}$, any AMEC of $\overline{\mathcal{P}}$ is the sub-graph of an AEMC of $\overline{\mathcal{R}}$.
\end{enumerate}
\end{lem}

Lemma~\ref{lem:Properties} can be verified by employing the proof of Theorem \ref{Thm_Properties}. Both of them demonstrate the advantages of applying the relaxed product MDP, which allows us to handle infeasible situations.

\begin{figure}[!t]\centering
	\subfloat[]{{\includegraphics[scale=0.46]{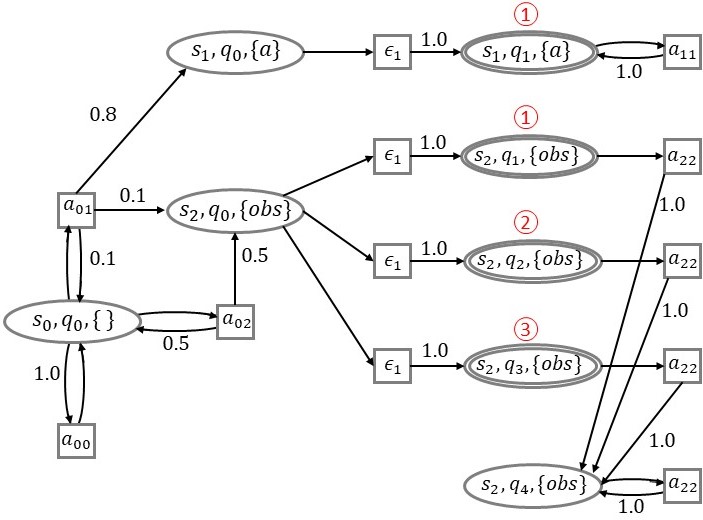} }}
	$\qquad$
	\subfloat[]{{
	\includegraphics[scale=0.39]{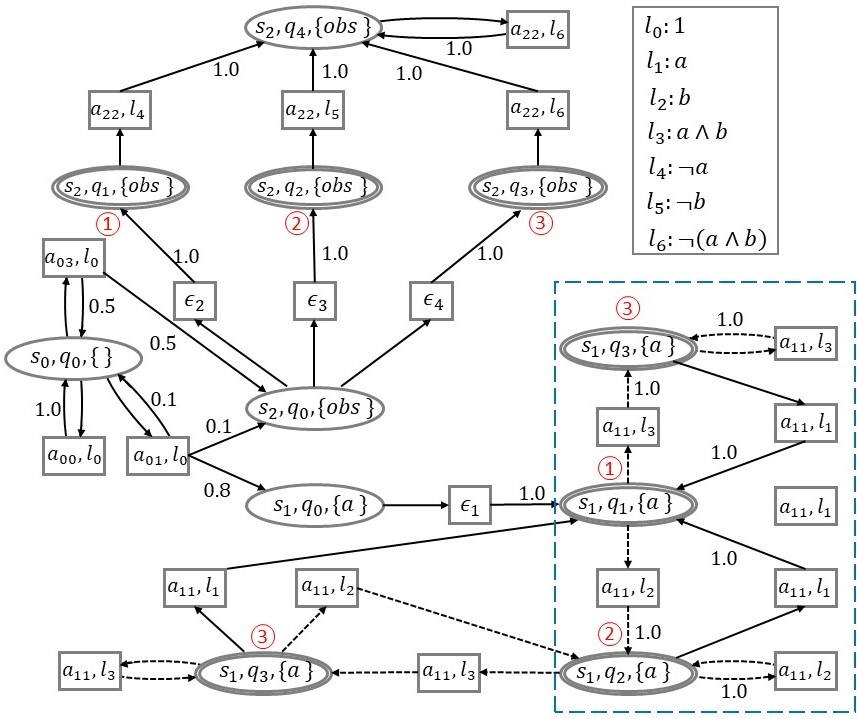}
		}}
	\caption{\label{fig:automaton_properties}  Two different product MDPs between the PL-MDP and the LDGBA provided in Fig.~\ref{fig:example1}. (a) Standard product MDP. There are no criteria to evaluate a policy for task satisfaction, and we can not find a path satisfying the acceptance condition.
	(b) Partial structure of the relaxed product MDP. $l_{i}, \forall i\in\left\{0,1,\ldots,7\right\}$ denotes the input alphabet of each automaton transition as shown in the gray rectangular. The gray dashed lines represent the transitions with nonzero violation costs, and the graph component in blue rectangular contains the path that satisfies the acceptance condition.}
\end{figure}

\begin{example}
 Fig.~\ref{fig:automaton_properties} (a) and (b) provide examples of a standard product MDP and the corresponding relaxed one, respectively, to illustrate the benefits of designing the relaxed product MDP. Both product MDPs are constructed between the PL-MDP  and the LDGBA of LTL formula $\phi=\oblong\lozenge\mathtt{a}\land\oblong\lozenge\mathtt{b}$, as shown in Fig.~\ref{fig:example1}. Obviously, the product MDP
in Fig.~\ref{fig:automaton_properties} (a) has no AMEC, while the relaxed product MDP in Fig.~\ref{fig:automaton_properties} (b) has one AMEC (blue rectangular) intersecting with all accepting sets. Moreover, the transitions with nonzero violation costs are gray dashed lines. Note that due to the complicated graph structure, we use the LDGBA in this example to illustrate the novelty of the proposed relaxed product MDP. Its advantages for infeasible cases are also applicable to the E-LDGBA.
\end{example}

Given a relaxed product MDP $\mathcal{\overline{\mathcal{R}}}$, let $MC_{\overline{\mathcal{R}}}^{\pi}$
denote the Markov chain induced by the policy $\pi$ on $\mathcal{\overline{\mathcal{R}}}$,
whose states can be represented by a disjoint union of a transient
class $\ensuremath{\ensuremath{T_{\pi}}}$ and $n$ closed irreducible
recurrent sets $R_{\pi}^{j}$, $j\in\left\{ 1,\ldots,n\right\} $,
i.e., $MC_{\overline{\mathcal{R}}}^{\pi}=\ensuremath{\ensuremath{T_{\pi}}}\sqcup\ensuremath{\ensuremath{\ensuremath{R_{\pi}^{1}}}\sqcup\ensuremath{\ensuremath{R_{\pi}^{2}}}\ldots\ensuremath{\ensuremath{R_{\pi}^{n}}}}$~\cite{Durrett1999}. 

\begin{lem}
\label{lemma:accepting set}Given a relaxed product MDP $\mathcal{\overline{\mathcal{R}}=M}\otimes\mathcal{\overline{A}}_{\phi}$
, the recurrent class $R_{\pi}^{j}$ of $MC_{\overline{\mathcal{R}}}^{\pi}$,
$\forall j\in\left\{ 1,\ldots,n\right\} $, induced by $\pi$ satisfies
one of the following conditions:
\begin{enumerate}
\item $\ensuremath{R_{\pi}^{j}}\cap F_{i}^{\overline{\mathcal{R}}}\neq\emptyset,\forall i\in\left\{ 1,\ldots f\right\} $,
or
\item $R_{\pi}^{j}\cap F_{i}^{\overline{\mathcal{R}}}=\emptyset,\forall i\in\left\{ 1,\ldots f\right\} $.
\end{enumerate}
\end{lem}

\begin{proof}
	The following proof is based on contradiction. Assume there exists
	a policy such that $\ensuremath{\ensuremath{R_{\pi}^{j}}}\cap F_{k}^{\overline{\mathcal{R}}}\neq\emptyset$,
	$\forall k\in K$, where $K$ is a subset of $2^{\left\{ 1,\ldots f\right\} }\setminus\left\{ \left\{ 1,\ldots f\right\} ,\emptyset\right\} $.
	As discussed in \cite{Durrett1999}, for each state in recurrent class,
	it holds that $\stackrel[n=0]{\infty}{\sum}p^{n}(x,x)=\infty$,
	where $x\in\ensuremath{\ensuremath{R_{\pi}^{j}}}\cap F_{k}^{\overline{\mathcal{R}}}$
	and $p^{n}(x,x)$ denotes the probability of returning
	from a transient state $x$ to itself in $n$ steps. This means that
	each state in the recurrent class occurs infinitely often. However,
	based on the embedded tracking frontier function of E-LDGBA in Def.
	\ref{def:E-LDGBA}, the tracking set $T$ will not be reset until
	all accepting sets have been visited. As a result, neither $\overline{q}_{k}\in \overline{F}_{k}$
	nor $x_{k}=(s,\overline{q}_{k})\in\ensuremath{\ensuremath{R_{\pi}^{j}}}\cap F_{k}^{\overline{\mathcal{R}}}$
	with $s\in S$ will occur infinitely, which contradicts the property
	$\stackrel[n=0]{\infty}{\sum}p^{n}(x_{k},x_{k})=\infty$.
\end{proof}

Lemma \ref{lemma:accepting set} indicates that, for any policy, all
accepting sets will be placed either in the transient class or the
recurrent class. As a result, the issue of NMEC, as in many existing
methods, can be avoided. Based on Theorem \ref{Thm_Properties} and
Lemma \ref{lemma:accepting set}, Problem \ref{Prob1} can be reformulated
as follows.
\begin{problem}
\label{Prob:2}Given a user-specified LTL task $\phi$
		and an MDP with unknown transition probabilities
		(i.e., motion uncertainties) and unknown labeling probabilities (i.e.,
		environment uncertainties), the objective is to find a policy in decreasing
		order of priority to 1) satisfy the acceptance condition of the relaxed
		product MDP, and 2) reduce the violation cost of the expected return.
\end{problem}

\section{Learning-Based Control Synthesis\label{sec:Solution}}

In this section, RL is leveraged
to identify policies for Problem~\ref{Prob:2}. Specifically, a model-free
multi-objective RL (MORL) is designed.

\subsection{Reward Design\label{subsec:Reward}}

The accepting reward function $\varLambda:X\shortrightarrow\mathbb{R}$
is designed as 
\begin{equation}
\varLambda(x)=\left\{ \begin{array}{cc}
r_{\varLambda} & \text{ if \ensuremath{\exists}i\ensuremath{\in\left\{  1,\ldots f\right\} } such that \ensuremath{x\in F_{i}^{\overline{\mathcal{R}}}} },\\
0 & \text{ otherwise},
\end{array}\right.\label{eq:AccRwd}
\end{equation}
where $r_{\varLambda}>0$. The violation function $V:X\times U^{\overline{\mathcal{R}}}\times X\shortrightarrow\mathbb{R}$
is designed as 
\begin{equation}
V(x,u^{\overline{\mathcal{R}}},x')=-c_{V}^{\overline{\mathcal{R}}}(x,u^{\overline{\mathcal{R}}},x').\label{eq:VioRwd}
\end{equation}

\sloppy The non-negative $\varLambda(x)$ enforces the accepting
condition of $\mathcal{\overline{\mathcal{R}}}$, while the non-positive function
$V(x,u^{\overline{\mathcal{R}}},x')$ indicates the penalty of violations.
Let $\boldsymbol{U}_{\boldsymbol{\pi}}=\left[\begin{array}{ccc}
U_{\boldsymbol{\pi}}(x_{0}) & U_{\boldsymbol{\pi}}(x_{1}) & \ldots\end{array}\right]^{T}\in\mathbb{R}^{N}$ denote the stacked expected return induced by $\boldsymbol{\pi}$
over $\mathcal{\overline{\mathcal{R}}}$ with $N=\left|X\right|$, the expected
return is designed as
\begin{equation}
\boldsymbol{U}=\stackrel[n=0]{\infty}{\sum}\gamma^{n}\boldsymbol{P}_{\boldsymbol{\pi}}^{n}(\boldsymbol{\varLambda}_{\boldsymbol{\pi}}+\beta\boldsymbol{V}_{\boldsymbol{\pi}}\cdot\mathbf{1}_{N}),\label{eq:expected_discount_return}
\end{equation}
where $0<\gamma<1$ is a discount factor, $\boldsymbol{P}_{\boldsymbol{\pi}}\in\mathbb{R}^{N\times N}$
is a matrix with entries representing the probabilities $p^{\overline{\mathcal{R}}}(x,\boldsymbol{\pi}(x),x')$
under $\pi$ for all $x,x'\in X$, $\boldsymbol{\varLambda}_{\boldsymbol{\pi}}=\left[\begin{array}{ccc}
\varLambda(x_{0}) & \varLambda(x_{1}) & \ldots\end{array}\right]^{T}\in\mathbb{R}^{N}$ is the stacked state rewards, $\beta\in\mathbb{R}^{+}$ is a weight
indicating the relative importance, $\mathbf{1}_{N}$ is an $N$-dimensional
vector of ones, and $\boldsymbol{V}_{\boldsymbol{\pi}}=\boldsymbol{P}_{\boldsymbol{\pi}}\circ\boldsymbol{V}\in\mathbb{R}^{N\times N}$
is the Hadamard product of $\boldsymbol{P}_{\boldsymbol{\pi}}$ and
$\boldsymbol{V}$, i.e., $\boldsymbol{V}_{\boldsymbol{\pi}}=$$\left[p^{\overline{\mathcal{R}}}(x,a_{\boldsymbol{\pi}}(x),x')\cdotp V(x,u^{\overline{\mathcal{R}}},x')\right]_{N\times N}$
and $a_{\boldsymbol{\pi}}(x)$ represents taking action
at $x$ from policy $\boldsymbol{\pi}$.

The objective is to identify a stationary policy $\pi^{*}$ that maximizes
the expected return
\begin{equation}
\pi^{*}=\underset{\pi}{\arg\max}\stackrel[n=0]{\infty}{\sum}\gamma^{n}\boldsymbol{P}_{\boldsymbol{\pi}}^{n}(\boldsymbol{\varLambda}_{\boldsymbol{\pi}}+\beta\boldsymbol{V}_{\boldsymbol{\pi}}\cdot\mathbf{1}_{N})\label{eq:RL_optimal_policy}
\end{equation}
$U_{\boldsymbol{\pi}}(x)\leq U_{\boldsymbol{\pi}^{*}}(x)$
for all $x\in X$ if $\pi^{*}$ in (\ref{eq:RL_optimal_policy}) is
optimal.

\begin{thm}
\label{thm:1}
Consider a relaxed
MDP product $\mathcal{\overline{\mathcal{R}}=M}\otimes\mathcal{\overline{A}}_{\phi}$.
If there exists a policy $\boldsymbol{\bar{\pi}}$
such that an induced run $r_{\overline{\mathcal{P}}}^{\boldsymbol{\bar{\pi}}}$
satisfies the acceptance condition of $\overline{\mathcal{R}}$, any optimization
method that solves~\eqref{eq:RL_optimal_policy} can find the policy
$\boldsymbol{\bar{\pi}}$.
\end{thm}

\begin{proof}
Detailed rigorous analysis can be found in~appendix\ref{appendix:A}
\end{proof}

To solve Problem \ref{Prob:2} by optimizing the expected return of \eqref{eq:expected_discount_return}, {$r_{\varLambda}$,
$\beta$ ,$\gamma$ can be determined as follows. Firstly, we can
choose a fixed $r_{\varLambda}$. Then, $\gamma$ can be obtained
by solving~\eqref{eq:case2_3} and~\eqref{eq:proof_case_2}. Finally, the range of $\beta$ can
be determined by solving~\eqref{eq:proof_case_1} and~\eqref{eq:proof_case_2}.
In order to minimize the violation cost, a great value of $\beta$
is preferred.

\begin{algorithm}
	\caption{\label{Alg3}  \textcolor{black}{Model-free RL-based control on MDPs under potentially soft LTL specifications.}}
	
	\singlespacing
	
	\begin{algorithmic}[1]
		
		\Procedure {Input: } {$\mathcal{M}$ , $\phi$, $\varLambda$}
		
		{Output: } { optimal policy $\boldsymbol{\pi}^{*}$ }
		
		{Initialization: } { Set $episode=0$ , $iteration=0$ and \newline\hspace*{8.0em} $\tau$ as maximum allowed learning steps }
		
		\State Initialize $r_{\varLambda}$, $\beta$ ,$\gamma$;
		
		\For { all $x\in X$ }
		
		\State $U(x)=0$ and $Q(x,u^{\mathcal{\mathcal{\mathcal{\mathcal{\overline{\mathcal{R}}}}}}})=0$
		$\forall x$, $u^{\mathcal{\mathcal{\mathcal{\mathcal{\overline{\mathcal{R}}}}}}}\in U^{\overline{\mathcal{R}}}(x)$
		
		\State $Count(x,u^{\mathcal{\mathcal{\mathcal{\mathcal{\overline{\mathcal{R}}}}}}})=0$
		for all $u^{\mathcal{\mathcal{\mathcal{\overline{\mathcal{R}}}}}}\in U^{\overline{\mathcal{R}}}(x)$
		
		\EndFor
		
		\State $x=x_{0}$;
		
		\While { $\boldsymbol{U}$ are not converged }
		
		\State $episode++$;
		
		\State $\epsilon=1/episode$;
		
		\While {$iteration<\tau$ }
		
		\State $iteration++$
		
		\State $u=\underset{u^{\overline{\mathcal{R}}}\in U^{\overline{\mathcal{R}}}}{\arg\max}Q(x,u^{\overline{\mathcal{R}}})$
		with probability  \newline
		\hspace*{4.2em} $1-\epsilon$, or select $u$ randomly from $U^{\overline{\mathcal{R}}}(x)$
		
		\State Obtain and execute $a$ of $\mathcal{M}$ from $u$
		
		\State Observe  $x'$, $\varLambda(x)$,$c_{V}^{\overline{\mathcal{R}}}(x,u,x')$

		\State $r\text{\ensuremath{\leftarrow}}\varLambda(x)-\beta c_{V}^{\overline{\mathcal{R}}}(x,u,x')$
		
		\State $Count(x,u)++$
		
		\State $\alpha=1/Count(x,u)$
		
		\State $Q(x,u)\text{\ensuremath{\leftarrow}}(1-\alpha)Q(x,u)+\newline
		\hspace*{4.6em}\alpha\left[r+\gamma\cdot\underset{u^{\overline{\mathcal{R}}}\in U^{\overline{\mathcal{R}}}}{\max}Q(x',u^{\overline{\mathcal{R}}})\right]$
		
		\State $x=x'$
		
		\EndWhile
		
		\EndWhile
		
		\For { all $x\in X$ }
		
		\State $U(x)=\underset{u^{\mathcal{\mathcal{\mathcal{\mathcal{\overline{\mathcal{R}}}}}}}\in U^{\mathcal{\mathcal{\overline{\mathcal{R}}}}}}{\max}Q(x,u^{\mathcal{\mathcal{\mathcal{\mathcal{\overline{\mathcal{R}}}}}}})$
		
		\State $\boldsymbol{\pi}^{*}(x)=\underset{u^{\mathcal{\mathcal{\mathcal{\mathcal{\overline{\mathcal{R}}}}}}}\in U^{\mathcal{\mathcal{\overline{\mathcal{R}}}}}}{\arg\max}U(x)$
		
		\EndFor
		
		\EndProcedure
		
	\end{algorithmic}
\end{algorithm}
	
\begin{rem}
Problem \ref{Prob:2} is a MORL problem, and Theorem \ref{thm:1} shows how it can be addressed to ensure  acceptance satisfaction while trying to minimize the long-term violation. In literature, MORL based
approaches often seek to identify Pareto fronts \cite{Vamplew2011}. 
However, our problem considers a trade-off between two
possibly conflicting objectives. There are no existing MORL methods to provide formal guarantee of both acceptance satisfaction and violation depreciation, which might result in a safety issue during the execution of optimal policies. To overcome this challenge, we can always divide the task into two parts: hard and soft constraints, with two parallel automatons.  The hard task should always be satisfied, and it can be applied with the traditional product MDP as Def.~\ref{def:P-MDP}. The soft part can be partially infeasible, and it can be relaxed as Def.~\ref{def:relaxed-product}. Such an idea can be found in our previous results~\cite{Cai2020receding}.
\end{rem} 

It is worth pointing out that $\boldsymbol{V}_{\pi}$ is a sparse
matrix because most transitions have zero violation cost, and Theorem
\ref{thm:1} provides a performance guarantee for the worst cases
of $\beta$ for acceptance satisfaction. By applying E-LDGBA, the sparse reward issue can
be improved compared with using LDGBA.

\subsection{Model-Free Reinforcement Learning~\label{subsec:RL}}

Q-learning is a model-free RL method~\cite{Watkins1992}, which can be used to find the optimal policy for a finite MDP. In particular, the agent updates its
Q-value from $x$ to $x'$ according to 
\begin{equation}
\begin{aligned}Q(x,u^{\overline{\mathcal{R}}})\text{\ensuremath{\leftarrow}} & (1-\alpha)Q(x,u^{\overline{\mathcal{R}}})\\
& +\alpha\left[R(x,u^{\overline{\mathcal{R}}},x')+\gamma\underset{u_{*}^{\overline{\mathcal{R}}}\in U^{\overline{\mathcal{R}}}(x)}{\max}Q(x,u_{*}^{\overline{\mathcal{R}}})\right],
\end{aligned}
\label{eq:Q}
\end{equation}
where $Q(x,u^{\overline{\mathcal{R}}})$ is the Q-value of the state-action
pair $(x,u^{\overline{\mathcal{R}}})$, $\alpha\in (0,1)$ is the learning
rate, $0<\gamma\leq1$ is the discount factor, and
\begin{equation}
R(x,u^{\overline{\mathcal{R}}},x')=\varLambda(x)+\beta V(x,u^{\overline{\mathcal{R}}},x')\label{eq:RL_reward}
\end{equation}
denotes the immediate reward from $x$ to $x'$ under $u^{\overline{\mathcal{R}}}$. The learning strategy is outlined in Alg. \ref{Alg3} that shows the steps of applying Q-learning into our framework. By applying the off-the-shelf RL, we have the following theorem.

\begin{thm}
\label{thm:convergence}
Given a finite MDP, i.e., $\overline{\mathcal{R}}=\mathcal{M}\otimes\mathcal{\overline{A}}_{\phi}=(X,U^{\overline{\mathcal{R}}},p^{\overline{\mathcal{R}}},x_{0},F^{\overline{\mathcal{R}}},c_{V}^{\overline{\mathcal{R}}},f_{V},T)$, let $Q^*(x,u^{\overline{\mathcal{R}}})$ be the optimal Q-function for every pair of state-action $(x,u^{\overline{\mathcal{R}}})$. Consider the RL with the updating rule

\begin{equation*}
\begin{aligned}Q_{k+1}(x,u^{\overline{\mathcal{R}}})\text{\ensuremath{\leftarrow}} & (1-\alpha_{k})Q_{k}(x,u^{\overline{\mathcal{R}}}) + \\
&\alpha_{k}\left[R(x,u^{\overline{\mathcal{R}}},x')+\gamma\underset{u_{*}^{\overline{\mathcal{R}}}\in U^{\overline{\mathcal{R}}}(x)}{\max}Q_{k}(x,u_{*}^{\overline{\mathcal{R}}})\right],
\end{aligned}
\end{equation*}
where $k$ is the updating step, $\gamma\in(0,1]$ is the discount factor, and the learning rate $\alpha_{k}$ satisfies $\underset{k}{\sum\alpha_{k}}=\infty$ and $\underset{k}{\sum\alpha^{2}_{k}}<\infty$. Then, $Q_{k}(x,u^{\overline{\mathcal{R}}})$  converges to $Q^*(x,u^{\overline{\mathcal{R}}})$ with a probability of $1$ as $k\rightarrow\infty$.
\end{thm}

\begin{figure}[t]
	\centering{}\includegraphics[scale=0.38]{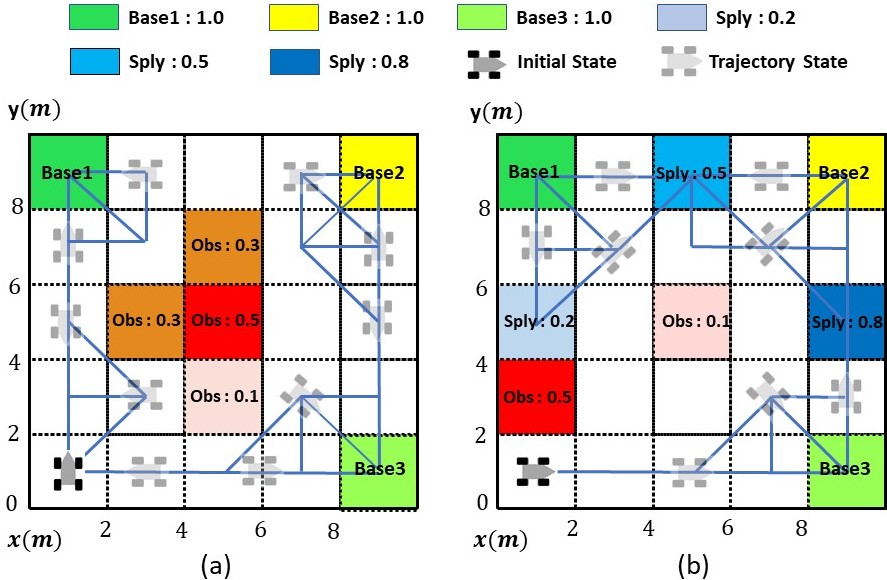}\caption{\label{fig:feasible_case} Simulated trajectories of 50 time steps under
		corresponding optimal policies in feasible workspaces with (a) a simple
		task, i.e., $\varphi_{case1}$, and (b) a relatively complex task, i.e., $\varphi_{case2}$.} 
\end{figure}

\begin{figure}[t]
	\centering{}\includegraphics[scale=0.30]{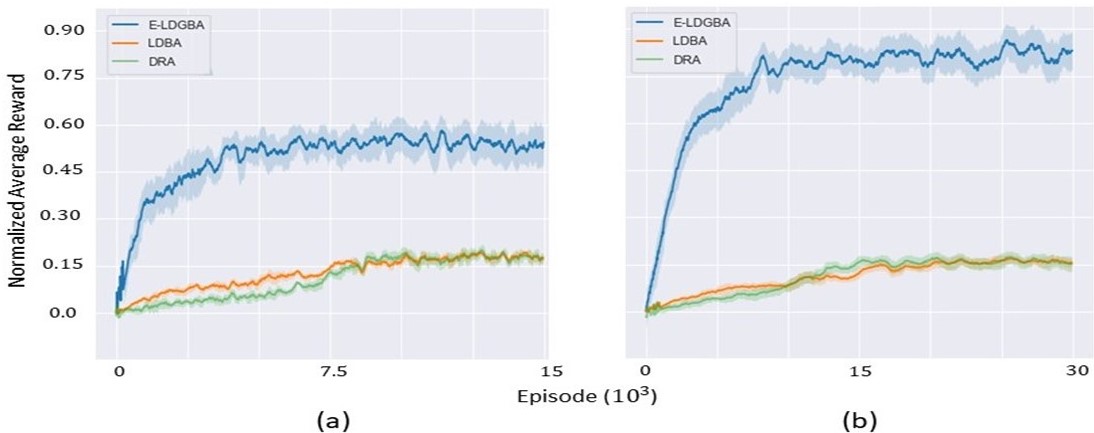}\caption{\label{fig:Feasible_Results} Normalized average rewards during training using E-LDGBA, LDBA, and DRA for (a) Task $\varphi_{case1}$ in Fig.~\ref{fig:feasible_case} and (b) Task $\varphi_{case2}$ in Fig.~\ref{fig:feasible_case}.} 
\end{figure}

Theorem~\ref{thm:convergence} is an immediate result of~\cite{tsitsiklis1994asynchronous}.
With standard learning rate $\alpha$ and discount factor $\gamma$
as Alg.~\ref{Alg3}, Q-value will converge to a unique limit
$Q^{*}$. Therefore, the optimal expected utility and policy can be
obtained as $U_{\boldsymbol{\pi}}^{*}(x)=\underset{u^{\overline{\mathcal{R}}}\in U^{\overline{\mathcal{R}}}(x)}{\max}Q^{*}(x,u^{\overline{\mathcal{R}}})$
and $\boldsymbol{\pi}^{*}(x)=\underset{u^{\overline{\mathcal{R}}}\in U^{\overline{\mathcal{R}}}(x)}{\arg\max}Q^{*}(x,u^{\overline{\mathcal{R}}})$.
In~(\ref{eq:Q}), the discount $\gamma$
is tuned to improve the trade-off between immediate and future rewards. Note that our novel design can be also applied with modern advanced RL algorithms.

\textbf{Complexity Analysis \text{ }} The number
of states is $\left|S\right|\times\left|Q\right|$, where $\left|Q\right|$
is determined by the original LDGBA $\mathcal{A}_{\phi}$ because
the construction of E-LDGBA $\mathcal{\overline{A}}_{\phi}$ will
not increase the size, and $\left|S\right|$ is the size of the labeled
MDP model. Due to the consideration of a relaxed product MDP and extended
actions in Def. \ref{def:relaxed-product}, the maximum complexity
of actions available at $x=(s,l,q)$ is $O(\left|A(s)\right|\times(\left|\Sigma\right|+1))$,
since $U^{\overline{\mathcal{R}}}(x)$ are created from $A(s)$
and $\Sigma\cup\left\{ \epsilon\right\} $.

\section{Case Studies\label{sec:Case}}
	
The developed RL-based control synthesis is implemented in Python.
Owl \cite{Kretinsky2018} is used to convert LTL specifications into
LDGBA, and P\_MDP package\cite{Guo2018} is used to construct state
transition models. All simulations are carried out on a laptop with
2.60 GHz quad-core CPU and 8 GB of RAM. For Q-learning, the optimal
policies of each case are generated using $10^{5}$ episodes with
random initial states. The learning rate $\alpha$ is determined by
Alg. \ref{Alg3} with $\gamma=0.999$ and $r_{\varLambda}=10$. To
validate the effectiveness of our approach, we first carry out simulations
over grid environments and then validate the approach in a more realistic office scenario with a TIAGo robot.

\subsection{Simulation Results}

\begin{figure}[t]
	\centering{}\includegraphics[scale=0.38]{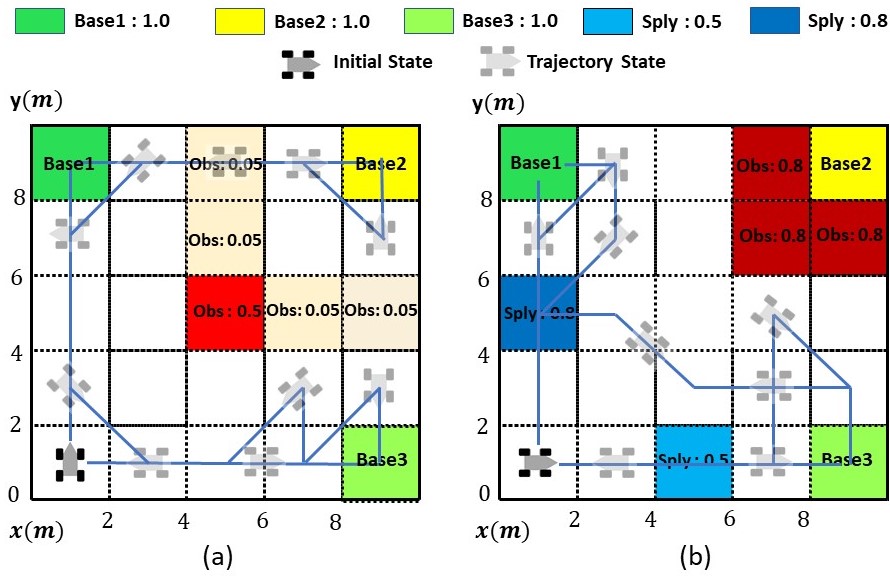}\caption{\label{fig:infeasible_case} Simulated trajectories of 50 time steps under corresponding optimal policies for partially infeasible tasks on workspaces where the agent visits the grid labeled as Base2 with (a) a low risk or (b) a high risk.} 
\end{figure}

\begin{figure}[t]
	\centering{}\includegraphics[scale=0.28]{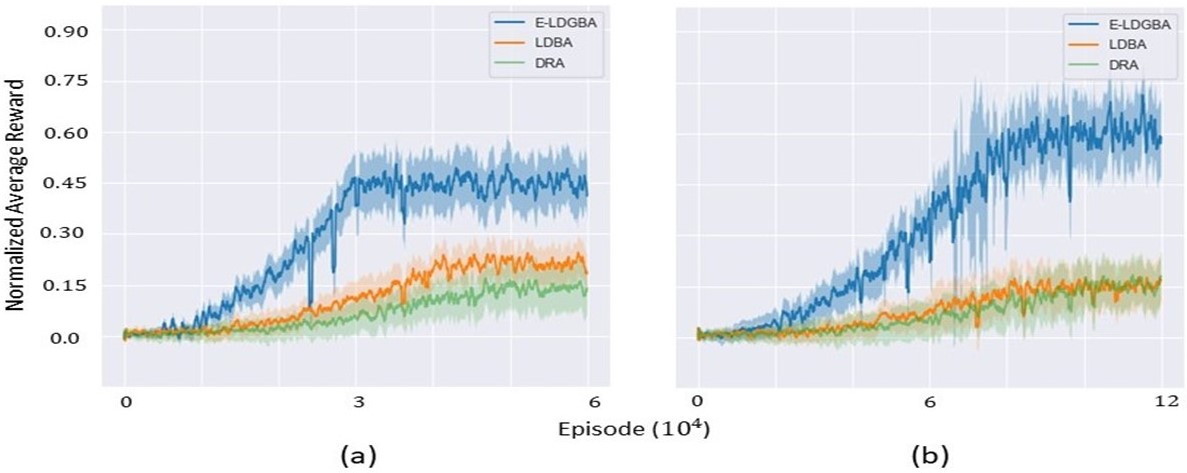}\caption{\label{fig:Infeasible_Results} Normalized average rewards during training using E-LDGBA, LDBA, and DRA for Task $\varphi_{case3}$ on two environments in Fig.~\ref{fig:infeasible_case}, respectively.} 
\end{figure}

Consider a partitioned $10m\times10m$ workspace, as shown in Fig.
\ref{fig:feasible_case} and Fig.
\ref{fig:infeasible_case}, where each cell is a $2m\times2m$ area. The
cells are marked with different colors to represent various areas
of interest, including $\mathtt{Base}1, \mathtt{Base}2, \mathtt{Base}3, \mathtt{Obs}$, and $\mathtt{Sply}$,
where $\mathtt{Obs}$ and $\mathtt{Sply}$ are shorthands for obstacle
and supply, respectively. To model environment uncertainties, the
number associated with a cell represents the likelihood that the corresponding
property appears at that cell. For example, $\mathtt{Obs}:0.1$ indicates
the obstacles occupy this cell with a probability of 0.1. The robot
dynamics follow the unicycle model, i.e., $\dot{x}=v\cdot\sin(\theta)$,
$\dot{y}=v\cdot\cos(\theta)$, and $\dot{\theta}=\omega$,
where $x,y,\theta$ indicate the robot positions and orientation. The linear and angular velocities are the control inputs, i.e.,
$u=(v,\omega)$. 
	
In addition, we assume the robot cannot consistently successfully execute the action primitives to model motion uncertainties. For instance, action primitives
``$FR$'' and ``$BK$'' mean the robot can successfully move forward
and backward $2m$ with a probability of $0.9$, respectively, and fail
with a probability of $0.1$. On the other hand, action primitives ``$TR$'' and ``$TL$''
mean the robot can successfully turn right and left for an angle of
$\frac{\pi}{2}$ exactly with a probability of $0.9$, respectively, and
fail by an angle of $\frac{\pi}{4}$ (undershoot) with a probability of
$0.05$ and an angle of $\frac{3\pi}{4}$ (overshoot) with another probability of
$0.05$. Fianlly, action primitive ``$ST$'' means the robot remains at its
current cell. The resulting MDP model has $25$ states. 
	
\textbf{Case 1: }As shown in Fig. \ref{fig:feasible_case} (a), we first
consider a case in which user-specified tasks can be successfully
executed. The desired surveillance task to be performed is formulated
as
\begin{equation}
\varphi_{case1}=(\oblong\lozenge\mathtt{Base}1)\land(\oblong\lozenge\mathtt{Base}2)\land(\oblong\lozenge\mathtt{Base}3)\land\oblong\lnot\mathtt{Obs},\label{eq:case1}
\end{equation}
which requires the mobile robot to visit all base stations infinitely
while avoiding obstacles. Its correspond\textcolor{black}{ing
	LDGBA has $2$ states with $3$ accepting sets}, and the relaxed product
MDP has $50$ states. In this case, each episode terminates after
$\tau=100$ steps and $\beta=8$. 
Fig. \ref{fig:feasible_case} (a) shows the generated optimal trajectory, which indicates $\varphi_{case1}$
is completed. 

\textbf{Case 2: }We validate our approach with more complex task specifications
in Fig. \ref{fig:feasible_case} (b). The task is expressed as
\begin{equation}
\varphi_{case2}=\varphi_{case1}\land\oblong\lozenge(\mathtt{Sply}\rightarrow\varbigcirc((\lnot\mathtt{Sply})\mathcal{U}\mathtt{\varphi_{one1}})),\label{eq:case2}
\end{equation}
where $\varphi_{one1}=\mathtt{Base}1\lor\mathtt{Base}2\lor\mathtt{Base}3$.
$\varphi_{case2}$ requires the robot to visit the supply station
and then go to one of the base stations while avoiding obstacles. It also requires all base stations to be visited. Its corresponding LDGBA
has $24$ states with $4$ accepting sets, and the relaxed product
MDP has $600$ states. The generated optimal trajectory is shown in
Fig. \ref{fig:feasible_case} (b).

\textbf{Case 3: }We consider more challenging workspaces in Fig. \ref{fig:infeasible_case}
(a) and (b), where user-specified tasks might not be fully feasible
due to potential obstacles (i.e., environment uncertainties). The
task specification is
\begin{equation}
\varphi_{case3}=\varphi_{case1}\land\oblong\lozenge(\varphi_{one1}\rightarrow\varbigcirc((\lnot\varphi_{one1})\mathcal{U}\mathtt{Sply})),\label{eq:case3}
\end{equation}
where $\varphi_{case3}$ requires the robot to visit one of the base
stations and then go to one of the supply stations while avoiding
obstacles. Its corresponding LDGBA has $24$ states with $4$ accepting
sets, and the product-MDP has $600$ states. 

For $\varphi_{case2}$
and $\varphi_{case3}$, each episode terminates after $\tau=1000$
steps and $\beta=5$. Note that in the case of Fig. \ref{fig:infeasible_case}
(a) and (b), AMECs might not exist since $\mathtt{Base}2$ is
surrounded by probabilistic obstacles and may not be visited. The simulated optimal
trajectories indicate that the robot takes a risk to accomplish the
task in the case of Fig. \ref{fig:infeasible_case} (a). In contrast, the robot
decides not to visit $\mathtt{Base}2$ to avoid the high risk of running
into obstacles in the case of Fig. \ref{fig:infeasible_case} (b). 

\textbf{Scalability Analysis: } The RL-based policy synthesis
is performed for $\varphi_{case1}$ over workspaces of various sizes (each grid is further partitioned) to show the computational complexity. The simulation results are listed
in Table \ref{tab:case1}. The number of episodes in Table \ref{tab:case1}
indicates the time used to converge to optimal satisfaction planing. It is also
verified that the given task $\varphi_{case1}$ can be successfully
carried out in large workspaces.

\begin{table}
    \caption{\label{tab:case1}Simulation results of large scale workspaces}
    \centering{}%
    \begin{tabular}{c|cccc}
        \hline 
        Workspace & MDP & $\overline{\mathcal{R}}$ & Episode & \tabularnewline
        size{[}cell{]} & States & States & Steps & \tabularnewline
        $15\times15$ & 225 & 450 & 30000 & \tabularnewline
        $25\times25$ & 625 & 1250 & 50000 & \tabularnewline
        $40\times40$ & 1600 & 3200 & 100000 & \tabularnewline
        \hline 
    \end{tabular}
\end{table}

\begin{figure}[!t]\centering
	\subfloat[]{{\includegraphics[scale=0.30]{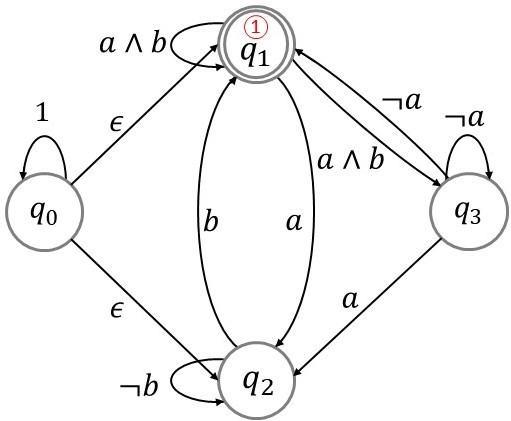} }}
	\subfloat[]{{\includegraphics[scale=0.30]{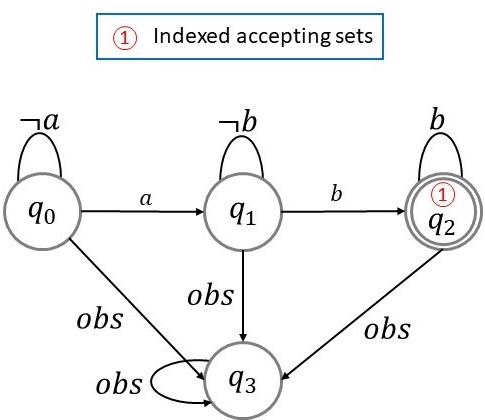} }}

	$\qquad$
	\subfloat[]{{
	\includegraphics[scale=0.44]{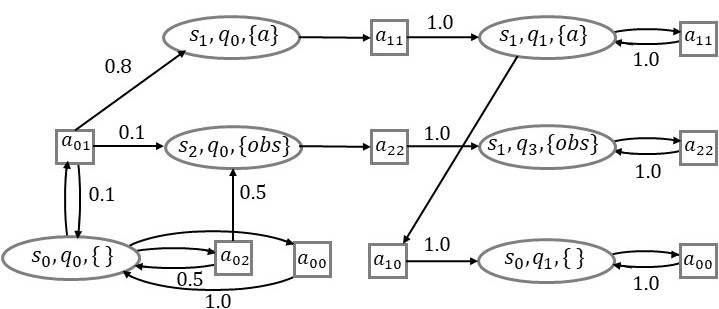}
		}}
    $\qquad$
	\subfloat[]{{
	\includegraphics[scale=0.44]{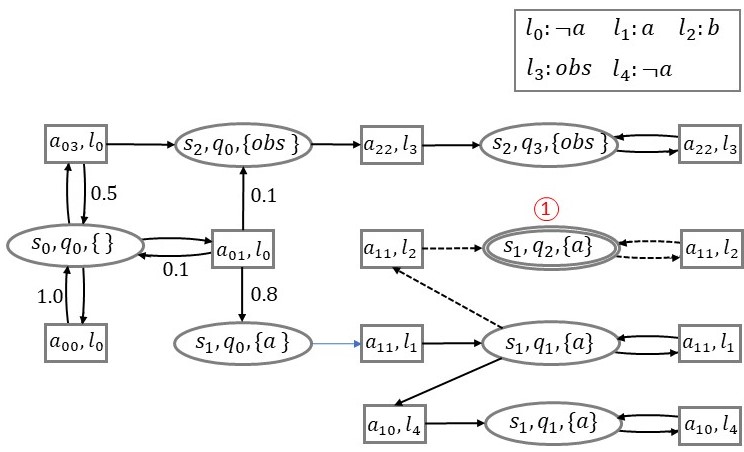}
		}}
	\caption{\label{fig:comparison}  Examples of the baseline comparisons regarding automaton structures and infeasible cases. (a) The LDBA of the LTL formula $\phi=\oblong\lozenge\mathtt{a}\land\oblong\lozenge\mathtt{b}$ only has one accepting set. (b) The LDGBA of the LTL formula $\phi=\lozenge(\mathtt{a}\land\lozenge\mathtt{b})$. (c) The standard product MDP between the PL-MDP in Fig.~\ref{fig:example1} (a) and the LDGBA of the LTL formula $\phi=\lozenge(\mathtt{a}\land\lozenge\mathtt{b})$ in Fig.~\ref{fig:comparison} (b). There are no accepting states, and this is an infeasible case. (d) The relaxed product MDP (Def.~\ref{def:relaxed-product}) corresponding to the standard product MDP in (c). An accepting state exists, and the dashed arrows represent the transitions with nonzero violation costs.
	}
\end{figure}

\subsection{Experimental Results}


\begin{figure}
	\centering{}\includegraphics[scale=0.28]{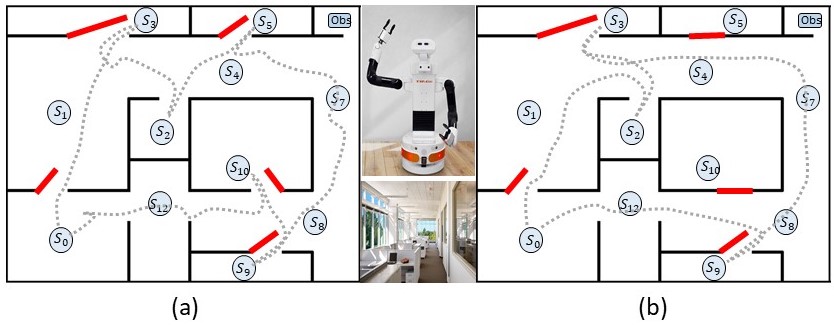}\caption{\label{fig:case_study3} \textcolor{black}{Experimental trajectories
			for the task $\varphi_{case4}$.}}
\end{figure}

This section verifies our algorithm for high-level decision-making problems in a real-world environment and shows that the framework can be adopted with low-level noisy controllers to formulate a hierarchical architecture.
Consider an office environment constructed in ROS Gazebo, as shown
in Fig~\ref{fig:case_study3}, which consists of $7$ rooms denoted
by $S_{0},S_{2},S_{3},S_{5},S_{9},S_{10},\mathtt{Obs}$ and $5$ corridors
denoted by $S_{1},S_{4},S_{7},S_{8},S_{12}$. The TIAGo robot can
follow a collision-free path from the center of one region to another
without crossing other regions using obstacle-avoidance navigation.
To model motion uncertainties, it is assumed that the robot can successfully
follow its navigation controller by moving to the desired region with a probability of
$0.9$ and fail by moving to an adjacent region with a probability of
$0.1$. The resulting MDP has $12$ states. The service to be performed
by TIAGo is expressed as
\begin{equation}
\varphi_{case4}=\varphi_{all}\land\oblong\lnot\mathtt{Obs}\label{eq:case4}
\end{equation}
wher\textcolor{black}{e $\varphi_{all}=\oblong\lozenge S_{0}\land\oblong\lozenge S_{2}\land\oblong\lozenge S_{3}\land\oblong\lozenge S_{5}\land\oblong\lozenge S_{9}\land\oblong\lozenge S_{10}$.
	In (\ref{eq:case4}), $\varphi_{all}$ requires the robot to always
	service all rooms (e.g., pick trash) and return to $S_{0}$ (e.g., release
	trash) while avoiding $\mathtt{Obs}$. Its corresponding LDGBA has
	$6$ states with $6$ accepting sets, and the relaxed product MDP has
	$72$ states.}

\textcolor{black}{All room doors are open, except the doors of room
	$S_{5}$ $S_{10}$ in Fig.~\ref{fig:case_study3} (b). As a result,
	$\varphi_{case4}$ can not be fully completed in the case of Fig.
	\ref{fig:case_study3} (b), and the task is infeasible. It is also worth
	pointing out there do not exist AMECs in the corresponding product automaton
	$\overline{\mathcal{P}}$ or $\mathcal{\overline{\mathcal{P}}}$ in Fig. \ref{fig:case_study3}
	(a) because the robot has the nonzero
	probability of entering $Obs$ at state $S_{7}$ as the motion uncertainties arise. The optimal policies
	for the two cases are generated, and each episode terminates after
	$\tau=150$ steps with $\beta=4$. The generated satisfying trajectories
	(without collision) of one round are shown in Fig. }\ref{fig:case_study3}
(a) and (b), and the robot tries to complete the feasible part of
task $\varphi_{case4}$ in Fig. \ref{fig:case_study3} (b).
	
\section{Conclusions}
	
This paper presents a learning-based control
synthesis of motion planning subject to motion and environment uncertainties. The LTL formulas are applied to express complex tasks via the automaton theory. Differently, in this work, an LTL formula is converted into a designed E-LDGBA to improve the performance
of mission accomplishment and resolve the drawbacks of DRA and LDBA. Furthermore, the innovative relaxed product MDP and utility schemes consisting of the accepting reward and violation reward are proposed to accomplish the satisfaction of soft tasks. 
In order to provide formal guarantees of achieving the goals of multi-objective optimizations, future research will consider more advanced
multi-objective learning methods. Also problems over continuous state and/or action spaces will be studied by incorporating deep neural networks.


\begin{appendices}

\section{Proof of Theorem~\ref{thm:1}}
\label{appendix:A}

For any policy $\boldsymbol{\pi}$, let $MC_{\overline{\mathcal{R}}}^{\boldsymbol{\pi}}$
denote the Markov chain induced by \textcolor{black}{$\pi$} on $\overline{\mathcal{R}}$.
Since$MC_{\overline{\mathcal{R}}}^{\boldsymbol{\pi}}$ can
    be written as $MC_{\overline{\mathcal{R}}}^{\boldsymbol{\pi}}=\ensuremath{\ensuremath{T_{\boldsymbol{\pi}}}}\sqcup\ensuremath{\ensuremath{\ensuremath{R_{\boldsymbol{\pi}}^{1}}}\sqcup\ensuremath{\ensuremath{R_{\boldsymbol{\pi}}^{2}}}\ldots\ensuremath{\ensuremath{R_{\boldsymbol{\pi}}^{n}}}}$,
(\ref{eq:expected_discount_return}) can be reformulated as 
\begin{equation}
\begin{aligned}\left[\begin{array}{c}
\boldsymbol{U}_{\boldsymbol{\pi}}^{tr}\\
\boldsymbol{U}_{\boldsymbol{\pi}}^{rec}
\end{array}\right]= & \stackrel[n=0]{\infty}{\sum}\gamma^{n}\left[\begin{array}{cc}
\boldsymbol{P}_{\pi}\left(\ensuremath{\mathcal{T}},\ensuremath{\mathcal{T}}\right) & \boldsymbol{P}_{\boldsymbol{\pi}}^{tr}\\
\boldsymbol{0}_{\sum_{i=1}^{m}N_{i}\times r} & \boldsymbol{P}_{\boldsymbol{\pi}}\left(R,R\right)
\end{array}\right]^{n}\\
& \cdot\left(\left[\begin{array}{c}
\boldsymbol{\varLambda}_{\boldsymbol{\pi}}^{tr}\\
\boldsymbol{\varLambda}_{\boldsymbol{\pi}}^{rec}
\end{array}\right]+\beta\left[\begin{array}{c}
\boldsymbol{\boldsymbol{O}}_{\boldsymbol{\pi}}^{tr}\\
\boldsymbol{\boldsymbol{O}}_{\boldsymbol{\pi}}^{rec}
\end{array}\right]\right),
\end{aligned}
\label{eq: expected_return_proof}
\end{equation}
where $\boldsymbol{U}_{\boldsymbol{\pi}}^{tr}$ and $\boldsymbol{U}_{\boldsymbol{\pi}}^{rec}$
are the utilities of states in the transient and recurrent classes,
respectively. In (\ref{eq: expected_return_proof}), $\boldsymbol{P}_{\boldsymbol{\pi}}\left(\ensuremath{\mathcal{T}},\ensuremath{\mathcal{T}}\right)\in\mathbb{R}^{r\times r}$
denotes the probability transition matrix between states in $\ensuremath{\ensuremath{T^{\boldsymbol{\pi}}}}$.
$\boldsymbol{P}_{\pi}^{tr}=\left[\boldsymbol{P}_{\pi}^{tr_{1}}\ldots\boldsymbol{P}_{\pi}^{tr_{m}}\right]\in\mathbb{R}^{r\times\sum_{i=1}^{m}N_{i}}$
is a probability transition matrix where $\boldsymbol{P}_{\pi}^{tr_{i}}\mathbb{\in R}^{r\times N_{i}}$
represents the probability of transiting from a transient state in
$\ensuremath{\ensuremath{T^{\pi}}}$ to the states of $\ensuremath{\ensuremath{R_{\pi}^{i}}}$.
The $\boldsymbol{P}_{\boldsymbol{\pi}}\left(R,R\right)$ is a diagonal
block matrix, where the $i$th block is an $N_{i}\times N_{i}$ matrix
containing transition probabilities between states within $\ensuremath{\ensuremath{R_{\boldsymbol{\pi}}^{j}}}$.
$\boldsymbol{P}^{\boldsymbol{\pi}}\left(\overline{\mathcal{R}},\overline{\mathcal{R}}\right)$
is a stochastic matrix since each block matrix is a stochastic matrix
\cite{Durrett1999}. The rewards vector $\boldsymbol{\varLambda}_{\boldsymbol{\pi}}$
can also be partitioned to $\boldsymbol{\varLambda}_{\boldsymbol{\pi}}^{tr}$
and $\boldsymbol{\varLambda}_{\boldsymbol{\pi}}^{rec}$. Similarly,
$\boldsymbol{\boldsymbol{O}}_{\boldsymbol{\pi}}=\left[\begin{array}{ccc}
O_{\boldsymbol{\pi}}\left(x_{0}\right) & O_{\boldsymbol{\pi}}\left(x_{1}\right) & \ldots\end{array}\right]^{T}=\boldsymbol{V}_{\boldsymbol{\pi}}\cdot\mathbf{1}_{N}$ can be divided into transient class $\boldsymbol{\boldsymbol{O}}_{\boldsymbol{\pi}}^{tr}$
and recurrent class $\boldsymbol{\boldsymbol{O}}_{\boldsymbol{\pi}}^{rec}$. 

We prove this theorem by contradiction. Suppose
there exists an optimal policy $\pi^{*}$ not satisfying the acceptance
condition of $\overline{\mathcal{R}}$. Based on Lemma \ref{lemma:accepting set},
the following is true: $F_{k}^{\overline{\mathcal{R}}}\subseteq\ensuremath{\ensuremath{\ensuremath{\ensuremath{T_{\boldsymbol{\pi}^{*}}}}}},\forall i\in\left\{ 1,\ldots f\right\} $.
As a result, for any $j\in\left\{ 1,\ldots n\right\} $, we have $\ensuremath{\ensuremath{R_{\boldsymbol{\pi}^{*}}^{j}}}\cap F_{i}^{\overline{\mathcal{R}}}\neq\emptyset,\forall i\in\left\{ 1,\ldots f\right\} $. 

The strategy is to show that there always exists
a policy $\bar{\pi}$ with greater utility than $\boldsymbol{\pi}^{*}$,
which contradicts to the optimality of $\pi^{*}$. Let's
consider a state $x_{R}\in\ensuremath{\ensuremath{R_{\boldsymbol{\pi}^{*}}^{j}}}$
and let $\boldsymbol{P}_{\boldsymbol{\pi}^{*}}^{x_{R}R_{\pi^{*}}^{j}}$
denote a row vector of $\boldsymbol{P}_{\boldsymbol{\pi}^{*}}^{n}\left(R,R\right)$
that contains the transition probabilities from $x_{R}$ to the states
in the same recurrent class $\ensuremath{\ensuremath{R_{\boldsymbol{\pi}^{*}}^{j}}}$
in $n$ steps. The expected return $U_{\boldsymbol{\pi}^{*}}\left(x_{R}\right)$
of $x_{R}$ under $\boldsymbol{\pi}^{*}$ is then obtained from (\ref{eq: expected_return_proof})
as: $U_{\boldsymbol{\pi}^{\boldsymbol{*}}}^{rec}\left(x_{R}\right)=\stackrel[n=0]{\infty}{\sum}\gamma^{n}\left[\boldsymbol{0}_{k_{1}}^{T}\:\boldsymbol{P}_{\boldsymbol{\pi}^{*}}^{x_{R}R_{\boldsymbol{\pi}^{*}}^{j}}\boldsymbol{0}_{k_{2}}^{T}\right]\left(\boldsymbol{\varLambda}_{\boldsymbol{\pi}^{*}}^{rec}+\beta\boldsymbol{O}_{\boldsymbol{\pi}^{*}}^{rec}\right)$
where $k_{1}=\sum_{i=1}^{j-1}N_{i}$, $k_{2}=\sum_{i=j+1}^{n}N_{i}$.
Since $\ensuremath{\overline{\mathcal{R}}_{\boldsymbol{\pi}^{*}}^{j}}\cap F_{i}^{\overline{\mathcal{R}}}=\emptyset,\forall i\in\left\{ 1,\ldots f\right\} $,
all the elements of $\boldsymbol{\varLambda}_{\boldsymbol{\pi}^{*}}^{rec}$
are equal to zero and each entry of $\boldsymbol{O}_{\pi^{*}}^{rec}$
is non-positive. We can conclude $U_{\pi^{*}}^{rec}\left(x_{R}\right)\leq0$.
To prove that optimal policy $\pi^{*}$ raises a contradiction, the
following analysis will show that $U_{\boldsymbol{\bar{\pi}}}^{rec}\left(x_{R}\right)\geq U_{\boldsymbol{\pi}^{*}}^{rec}\left(x_{R}\right)$
for some policies $\boldsymbol{\bar{\pi}}$ that satisfy the acceptance
condition of $\mathcal{\overline{\mathcal{R}}}$. Thus we have $\ensuremath{R_{\boldsymbol{\bar{\pi}}}^{j}}\cap F_{i}^{\overline{\mathcal{R}}}\neq\emptyset,\forall i\in\left\{ 1,\ldots f\right\} $.

\sloppy
\textbf{Case 1:} If $x_{R}\in\ensuremath{R_{\boldsymbol{\bar{\pi}}}^{j}}$,
by Lemma \ref{lemma:accepting set} and (\ref{eq:AccRwd}), there
exist a minimum of $f$ accepting states such that $X_{\varLambda}=\left\{ x_{\varLambda}\left|x_{\varLambda}\in\ensuremath{R_{\boldsymbol{\bar{\pi}}}^{j}}\cap F_{i}^{\overline{\mathcal{R}}},\forall i\in\left\{ 1,\ldots f\right\} )\right.\right\} $
in $\ensuremath{R_{\boldsymbol{\bar{\pi}}}^{j}}$ with positive rewards
$r_{\varLambda}$. From (\ref{eq: expected_return_proof}), $U_{\boldsymbol{\bar{\pi}}}^{rec}\left(x_{R}\right)$
can be lower bounded as $U_{\boldsymbol{\bar{\pi}}}^{rec}\left(x_{R}\right)\geq\stackrel[n=0]{\infty}{\sum}\gamma^{n}\left(P_{\boldsymbol{\bar{\pi}}}^{x_{R}x_{\varLambda}}r_{\varLambda}+\beta\boldsymbol{P}_{\boldsymbol{\bar{\pi}}}^{x_{R}\ensuremath{R_{\bar{\pi}}^{j}}}\boldsymbol{V}_{\boldsymbol{\bar{\pi}}}^{rec}\mathbf{1}_{N_{j}}\right),$
where $P_{\bar{\pi}}^{x_{R}x_{\varLambda}}$ is the transition probability
from $x_{R}$ to $x_{\varLambda}$ in $n$ steps, and $\boldsymbol{V}_{\boldsymbol{\bar{\pi}}}^{rec}\in\mathbb{R}^{N_{j}\times N_{j}}$
represents the violation cost of states in $\overline{\mathcal{R}}_{\boldsymbol{\bar{\pi}}}^{j}$.
Since $x_{R}$ and $x_{\varLambda}$ are recurrent states, there always
exists a lower bound $\underline{P}_{\boldsymbol{\bar{\pi}}}^{x_{R}x_{\varLambda}}\in\mathbb{R}^{+}$
of the transition probability $P_{\boldsymbol{\bar{\pi}}}^{x_{R}x_{\varLambda}}$.
We can select a positive reward $r_{\varLambda}$ such that
\begin{equation}
U_{\boldsymbol{\bar{\pi}}}^{rec}\left(x_{R}\right)\geq\underline{P}_{\boldsymbol{\bar{\pi}}}^{x_{R}x_{\varLambda}}r_{\varLambda}+\beta N_{j}^{2}\underline{V}_{\boldsymbol{\bar{\pi}}}^{rec}\geq0\label{eq:proof_case_1}
\end{equation}
where $\underline{V}_{\boldsymbol{\bar{\pi}}}^{rec}\in\mathbb{R}^{-}$
represents the minimal entry in $\boldsymbol{V}_{\boldsymbol{\bar{\pi}}}^{rec}$.
By selecting $r_{\varLambda}$ to satisfy (\ref{eq:proof_case_1}),
we can conclude in this case $U_{\boldsymbol{\bar{\pi}}}^{rec}\left(x_{R}\right)>U_{\boldsymbol{\pi}^{*}}^{rec}\left(x_{R}\right)$.

\textbf{Case 2:}.If $x_{R}\in\mathcal{T}_{\boldsymbol{\bar{\pi}}}$,
we know $\mathcal{T}_{\boldsymbol{\bar{\pi}}}\cap F_{i}^{\overline{\mathcal{R}}}=\emptyset,\forall i\in\left\{ 1,\ldots f\right\} $.
As demonstrated in \cite{Durrett1999}, for a transient state $x_{tr}\in\mathcal{T}_{\boldsymbol{\bar{\pi}}}$,
there always exists an upper bound $\Delta<\infty$ such that $\stackrel[n=0]{\infty}{\sum}p^{n}\left(x_{tr},x_{tr}\right)<\Delta$,
where $p^{n}\left(x_{tr},x_{tr}\right)$ denotes the probability of
returning from a transient state $x_{tr}$ to itself in $n$ time
steps. For a recurrent state $x_{rec}\in\overline{\mathcal{R}}_{\boldsymbol{\bar{\pi}}}^{j}$,
it is always true that 
\begin{equation}
\stackrel[n=0]{\infty}{\sum}\gamma^{n}p^{n}\left(x_{rec},x_{rec}\right)>\frac{1}{1-\gamma^{\overline{n}}}\bar{p}\label{eq:case2_1}
\end{equation}
where there exists $\overline{n}$ such that $p^{\overline{n}}\left(x_{rec},x_{rec}\right)$
is nonzero and can be lower bounded by $\bar{p}$ \cite{Durrett1999}.
From (\ref{eq: expected_return_proof}), one has 
\begin{equation}
\begin{aligned}\boldsymbol{U}_{\bar{\pi}}^{tr}> & \stackrel[n=0]{\infty}{\sum}\gamma^{n}\boldsymbol{P}_{\boldsymbol{\bar{\pi}}}^{n}\left(\ensuremath{\mathcal{T}},\ensuremath{\mathcal{T}}\right)\left(\boldsymbol{\varLambda}_{\bar{\pi}}^{tr}+\beta\boldsymbol{O}_{\bar{\pi}}^{tr}\right)\\
& +\stackrel[n=0]{\infty}{\sum}\gamma^{n}\boldsymbol{P}_{\boldsymbol{\bar{\pi}}}^{tr}\boldsymbol{P}_{\boldsymbol{\bar{\pi}}}^{n}\left(\overline{\mathcal{R}},\overline{\mathcal{R}}\right)\left(\boldsymbol{\varLambda}^{tec}+\beta\boldsymbol{O}_{\bar{\pi}}^{rec}\right)
\end{aligned}
\label{eq:case2_2}
\end{equation}

\sloppy Let $\max\left(\cdot\right)$ and $\min\left(\cdot\right)$ represent
the maximum and minimum entry of an input vector, respectively. The
upper bound $\bar{m}=\left\{ \max\left(\overline{M}\right)\left|\overline{M}<\boldsymbol{P}_{\bar{\pi}}^{tr}\boldsymbol{\bar{P}}\left(\boldsymbol{\varLambda}_{\bar{\pi}}^{rec}+\beta\boldsymbol{V}_{\bar{\pi}}^{rec}\mathbf{1}_{\tilde{N}}\right)\right.\right\} $,
where $\tilde{N}=\sum_{j=1}^{m}N_{j}$ and $\boldsymbol{\bar{P}}$
is a block matrix whose nonzero entries are derived similarly to the
$\bar{p}$ in (\ref{eq:case2_1}). \textcolor{black}{Using the fact
    $\stackrel[n=0]{\infty}{\sum}\gamma^{n}\boldsymbol{P}_{\boldsymbol{\bar{\pi}}}^{n}\left(\ensuremath{\mathcal{T}},\ensuremath{\mathcal{T}}\right)\leq\Delta\boldsymbol{1}_{r\times r}$}
\cite{Durrett1999}, where $\boldsymbol{1}_{r\times r}$
    is a $r\times r$ matrix of all ones, the utility $U_{\boldsymbol{\bar{\pi}}}^{tr}\left(x_{R}\right)$
can be lower bounded from (\ref{eq:case2_1}) and (\ref{eq:case2_2})
as

\begin{equation}
U_{\boldsymbol{\bar{\pi}}}^{tr}\left(x_{R}\right)>\Delta\cdot r\cdot\beta\underline{V}_{\bar{\pi}}^{tr}+\frac{1}{1-\gamma^{\overline{n}}}\bar{m}\label{eq:case2_3}
\end{equation}

Since $U_{\boldsymbol{\pi}^{*}}^{rec}\left(x_{R}\right)=0$, the contradiction
$U_{\boldsymbol{\bar{\pi}}}^{tr}\left(x_{R}\right)>0$ in this case
will be achieved if $\Delta\cdot r\cdot\beta\underline{V}_{\boldsymbol{\bar{\pi}}}^{tr}+\frac{1}{1-\gamma^{\overline{n}}}\bar{m}\geq0$.
Because $\Delta r\left(w_{L}+\beta\underline{V}_{\boldsymbol{\bar{\pi}}}^{tr}\right)<0$,
it needs $\bar{m}>0$ as

\begin{equation}
\bar{m}>r_{\varLambda}+\beta\tilde{N}\underline{V}_{\boldsymbol{\bar{\pi}}}^{rec}>0.\label{eq:proof_case_2}
\end{equation}

Thus, by choosing $\gamma$ to be sufficiently
close to $1$ with $\bar{m}>0$, we have $U_{\boldsymbol{\bar{\pi}}}^{tr}\left(x_{R}\right)>0\geq U_{\boldsymbol{\pi}^{*}}^{rec}\left(x_{R}\right)$.
The above procedure shows the contradiction of the assumption that
$\boldsymbol{\pi}_{V}^{*}$ is optimal.

\end{appendices}

\bibliographystyle{IEEEtran}
\bibliography{BibMaster}

\end{document}